\documentclass{bmvc2k}

\usepackage{verbatim}
\usepackage{amsmath}
\usepackage{amssymb}
\usepackage{bm}
\usepackage{booktabs} 
\usepackage{multirow}
\usepackage{graphicx}
\usepackage{wrapfig}
\usepackage{colortbl}  
\usepackage{xcolor}
\usepackage{array}   
\usepackage{wrapfig}
\newif\ifSubmitting
  \Submittingtrue 

\newif\ifPreparing
  \Preparingfalse

\newif\ifSecondPreparing
  \SecondPreparingfalse

\newif\ifThirdPreparing
  \ThirdPreparingtrue 

\ifSubmitting 
    \Preparingfalse
    \SecondPreparingfalse
    \ThirdPreparingfalse
\fi
\usepackage{soul}  
\usepackage{color}
\usepackage{xcolor}

\def\figname{Fig\bmvaOneDot}
\def\tablename{Table~}
\def\eqname{Eq\bmvaOneDot}
\def\secname{Sec\bmvaOneDot}

\makeatletter
\DeclareRobustCommand\onedot{\futurelet\@let@token\@onedot}
\def\@onedot{\ifx\@let@token.\else.\null\fi\xspace}

\def\etal{\emph{et al}\onedot}
\makeatother

\usepackage{pifont}
\newcommand{\cmark}{\ding{51}}%
\newcommand{\xmark}{\ding{55}}%

\newcommand{\jlnote}[2]{{\jlred [#1]}{{\jlgray #2}}}
\newcommand{\jlsecnote}[2]{{\jlsecred [#1]}{{\jlsecgray #2}}}
\newcommand{\jlthirdnote}[2]{{\jlthirdred [#1]}{{\jlthirdgray #2}}}

\newcommand{\jldelete}[1]{{\jlred\st{#1}}}
\newcommand{\jlsecdelete}[1]{{\jlsecred\st{#1}}}
\newcommand{\jlthirddelete}[1]{{\jlthirdred\st{#1}}}

\ifPreparing
    \newenvironment{jlblack}{\color{black}}{}
    \newenvironment{jlblue}{\color{blue}}{}
    \newenvironment{jlred}{\color{red}}{}
    \newenvironment{jlmagenta}{\color{magenta}}{}
    \newenvironment{jlorange}{\color{orange}}{}
    \newenvironment{jltan}{\color{tan}}{}
    \newenvironment{jlgray}{\color{gray}}{}
    \newenvironment{jllightgray}{\color{lightgray}}{}
\else
    
    \newenvironment{jlblue}{}{}
    \newenvironment{jlred}{}{}
    \newenvironment{jlmagenta}{}{}

    \newenvironment{jlgray}{}{}
    
    \renewcommand{\jlnote}[2]{{}{#2}}
    \renewcommand{\jldelete}[1]{}   
\fi
    \newcommand{\jlreplace}[2]{{\jlblue #2}}
    \newcommand{\jlyetreplace}[2]{{\jlmagenta #1}}
    \newcommand{\jlnoterem}[1]{}
    \newcommand{\jlcleandelete}[1]{}
    \newcommand{\jlinsert}[1]{{\jlblue #1}}

\ifSecondPreparing
    \newenvironment{jlsecblack}{\color{black}}{}
    \newenvironment{jlsecblue}{\color{blue}}{}
    \newenvironment{jlsecred}{\color{red}}{}
    \newenvironment{jlsecmagenta}{\color{magenta}}{}
    \newenvironment{jlsecorange}{\color{orange}}{}
    \newenvironment{jlsectan}{\color{tan}}{}
    \newenvironment{jlsecgray}{\color{gray}}{}
    \newenvironment{jlseclightgray}{\color{lightgray}}{}
\else
    
    \newenvironment{jlsecblue}{}{}
    \newenvironment{jlsecred}{}{}
    \newenvironment{jlsecmagenta}{}{}

    \newenvironment{jlsecgray}{}{}
    
    \renewcommand{\jlsecnote}[2]{{}{#2}}
    \renewcommand{\jlsecdelete}[1]{}    
\fi
    \newcommand{\jlsecreplace}[2]{{\jlsecblue #2}}

    \newcommand{\jlsecnoterem}[1]{}
    \newcommand{\jlseccleandelete}[1]{}

\ifThirdPreparing
    \newenvironment{jlthirdblack}{\color{black}}{}
    \newenvironment{jlthirdblue}{\color{blue}}{}
    \newenvironment{jlthirdred}{\color{red}}{}
    \newenvironment{jlthirdmagenta}{\color{magenta}}{}
    \newenvironment{jlthirdorange}{\color{orange}}{}
    \newenvironment{jlthirdtan}{\color{tan}}{}
    \newenvironment{jlthirdgray}{\color{gray}}{}
    \newenvironment{jlthirdlightgray}{\color{lightgray}}{}
\else
    
    \newenvironment{jlthirdblue}{}{}
    \newenvironment{jlthirdred}{}{}
    \newenvironment{jlthirdmagenta}{}{}

    \newenvironment{jlthirdgray}{}{}
    
    \renewcommand{\jlthirdnote}[2]{{}{#2}}
    \renewcommand{\jlthirddelete}[1]{}    
\fi
    \newcommand{\jlthirdreplace}[2]{{\jlthirdblue #2}}

    \newcommand{\jlthirdnoterem}[1]{}
    \newcommand{\jlthirdcleandelete}[1]{}
    \newcommand{\jlthirdinsert}[1]{{\jlthirdblue #1}}

\title{Learning Anatomically Consistent Embedding for Chest Radiography}

\addauthor{Ziyu Zhou*}{zhouziyu@sjtu.edu.cn}{1,2}
\addauthor{Haozhe Luo*}{hluo54@asu.edu}{2}
\addauthor{Jiaxuan Pang}{jpang12@asu.edu}{2}
\addauthor{Xiaowei Ding$\dagger$}{dingxiaowei@sjtu.edu.cn}{1}
\addauthor{Michael Gotway}{gotway.michael@mayo.edu}{3}
\addauthor{Jianming Liang$\dagger$}{Jianming.Liang@asu.edu}{2}
\addinstitution{
Shanghai Jiao Tong University, China
}
\addinstitution{
Arizona State University, USA
}
\addinstitution{
Mayo Clinic, USA
}

\runninghead{Z. Zhou, H. Luo et al.}{Learning Anatomically Consistent Embedding}

\def\etal{\emph{et al}\bmvaOneDot}

\begin{document}
\maketitle

\begin{abstract}
\jlnoterem{Wrting is wordy and loose, and you may want to add 1-2 sentences to summarize why the method works intuitively and the specific performance in terms of numbers.}{}
\noindent Self-supervised learning (SSL) approaches have recently shown substantial success in learning visual representations from unannotated images. Compared with photographic images, medical \jlreplace{image samples}{images acquired with the same imaging protocol} exhibit \jlreplace{highly consistent anatomy}{high consistency in anatomy}.\jlreplace{
    We introduce a novel SSL approach called PEAC for medical image analysis, \jlreplace{which effectively utilizes}{aiming to exploit the anatomical consistency with} both global and local structures within medical images.
}{
    To exploit this anatomical consistency\jlcleandelete{with both global and local structures within medical images}, this paper introduces a novel SSL approach, called PEAC (patch embedding of anatomical consistency), for medical image analysis.
}\jlreplace{Additionally}{Specifically, in this paper}, we propose to learn global and local consistencies via \jlsecnote{If you like this name, please include and explain it in the method section}{stable grid-based matching}\jlreplace{. We}{,} transfer pre-trained PEAC models to diverse downstream tasks\jlreplace{and generate visualizations. The experimental results demonstrate}{, and extensively demonstrate} that (1) PEAC achieves significantly better performance \jlreplace{compared to}{than the} existing state-of-the-art \jlreplace{contrastive learning}{fully/self-supervised} methods, and (2) PEAC\jlthirdcleandelete{ effectively} captures the anatomical structure consistency \jlthirdreplace{between}{across} \ul{views} of the same patient and \jlthirdreplace{between}{across} patients of \ul{different genders, weights, and healthy statuses}, \jlnoterem{I am not sure if we can claim this}{which enhances the interpretability of our method for medical image analysis.} 
\jlinsert{All code and pretrained models are available at \href{https://github.com/JLiangLab/PEAC}{GitHub.com/JLiangLab/PEAC}}.  
\end{abstract}

\section{Introduction}
\jlsecnote{Please develop a minimal set of consistent, coherent, and formal terminologies and use them in figures and text throughout the paper. I suggest that you document them in terminology.tex}{}

\begin{figure}[h!]
    \centering
    \setlength{\belowcaptionskip}{0.1cm}
    \includegraphics[width=10.5cm]{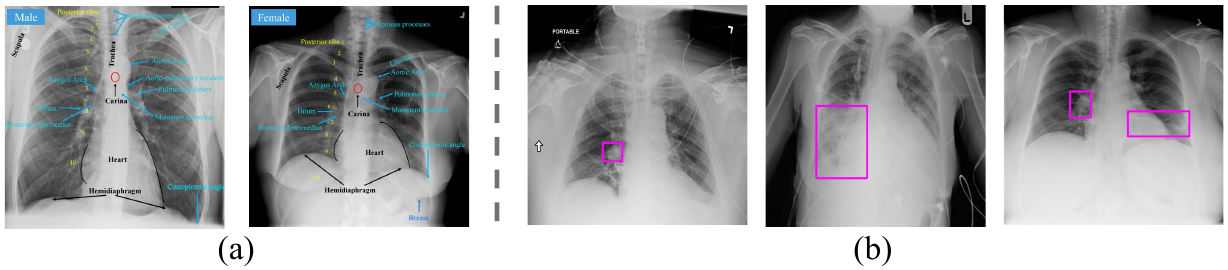}
    \caption{\jlthirdreplace{\jlnoterem{I would not advise using pink for those anatomical structures}{}\jlnote{writing is wordy and loose}{(a) Chest X-rays contain various large (global) and small (local) anatomical patterns that can be utilized for structural consistency, including the right/left lung, heart, spinous processes, clavicle, mainstem bronchus, hemidiaphragm, and the osseous structures of the thorax. (b) The diagnosis of chest diseases often involves identifying focal and diffuse patterns, such as \jlsecnoterem{please consider including X-rays for the focal and diffuse diseases in the lung}{Mass, Infiltrate, and Atelectasis} \jlreplace{within boxed areas of}{as boxed in} the images.}}
{(a) Chest X-rays contain various large (global) and small (local) anatomical patterns, including the right/left lung, heart, spinous processes, clavicle, mainstem bronchus, hemidiaphragm, and the osseous structures of the thorax, that can be utilized for learning consistent embedding in anatomy. (b) Diagnosing chest diseases at chest X-rays involves identifying focal and diffuse patterns, such as Mass, Infiltrate, and Atelectasis as boxed, that can be exploited for learning consistent embedding in disease. 
}}
    \label{fig:motivation}
\end{figure}

\jlreplace{
    \jlnote{the first two paragraphs do not align well yet. consider revisions for one coherent introduction.}

\label{sec:intro}

\jlyetreplace{
    Self-supervised learning (SSL) has been widely proven as an effective method for extracting image representations. It utilizes the image itself as supervised information such as context-instance relationships~\cite{doersch2015unsupervised}, instance-instance contrast~\cite{chen2020simple} or dense contrast~\cite{o2020unsupervised}. The research on SSL is particularly valuable for medical images since labeling medical images is expensive, time-consuming and requires experienced physicians~\cite{taleb2021multimodal}, and more importantly medical images are more challenging to obtain compared to natural images because of privacy concerns~\cite{tajbakhsh2021guest}. Many SSL methods have achieved good results on photographic images~\cite{xie2022simmim, xiao2021region}, but directly transfer to medical images can pose some problems. There might be several reasons. Firstly, the methods based on natural images focus more on foreground objects, while medical images require consideration of not only important foreground objects such as spine, bones, and tumors, but also equally important background structures such as soft tissues, blood vessels, and nerves. Secondly, photographic images like ImageNet~\cite{deng2009imagenet} often hold significant difference in background and a definite object in the center like a dog or a cat. Differently, medical images acquired with the same imaging protocol contain the same body part with similar structures and backgrounds. 
    %
    }{
    Self-supervised learning (SSL) has proven to be effective in learning image representation from unlabeled images by utilizing information embedded in images themselves as supervision signals, including context-instance relationships~\cite{doersch2015unsupervised}, instance-instance contrast~\cite{chen2020simple} or dense contrast~\cite{o2020unsupervised}, achieving impressive results in photographic images in computer vision~\cite{xie2022simmim,xiao2021region}. SSL is naturally more attractive in medical imaging because labeling medical images is more expensive and time-consuming and requires experienced physicians~\cite{zhou2017fine,taleb2021multimodal}, and privacy concerns make collecting medical images even more challenging~\cite{tajbakhsh2021guest}. Nevertheless, directly transferring these SSL developed for photographic images to medical images may not achieve optimal results because photographic images like those in ImageNet~\cite{deng2009imagenet} often have foreground objects (e.g., dogs and cats) at the center with backgrounds of large variation. As a result, the SSL methods developed for photographic images mostly focus on foreground objects. By contrast, medical images acquired with the same imaging protocol have similar anatomical structures, and imaging diagnosis requires considering not only ``foreground'' objects: diseases (abnormalities) but also ``background'': anatomy. Therefore, we hypothesize that SSL can offer better performance by learning from anatomy in health and disease.
    }
}{
    Self-supervised learning (SSL)~\citep{jing2020self} pretrains generic source models without using expert annotation, allowing the pretrained generic source models to be quickly fine-tuned into high-performance application-specific target models~\citep{zhou2021models} and minimizing annotation cost~\citep{Tajbakhsh2021a}. This paradigm is particularly attractive in medical imaging because labeling medical images is tedious, laborious, and time-consuming and demands specialty-oriented expertise~\cite{zhou2017fine,taleb2021multimodal}. However, most existing SSL methods were developed for photographic images, and directly adopting these SSL methods to medical images may not achieve optimal results because medical images are markedly different from photographic images. Photographic images, like those in ImageNet~\cite{deng2009imagenet}, are object-centric, where dominant objects (e.g., dogs and cats) are located at the center with backgrounds of large variation. Naturally, these SSL methods developed for photographic images mostly learn from foreground objects. By contrast, medical images acquired with the same imaging protocol have similar anatomical structures, and imaging diagnosis requires not only analyzing ``foreground'' objects---diseases, but also understanding ``background'' objects---anatomical structures. Furthermore, diseases are often small and obscured in ``background'' anatomical structures. Therefore, we hypothesize that SSL achieves better performance in medical imaging when learning from anatomy in health and disease.
To test this hypothesis, we have chosen chest X-rays because the chest contains several critical organs prone to a number of diseases associated with significant healthcare costs~\cite{zhou2021models}, and chest X-rays are one of the most frequently used modalities in imaging the chest. In chest X-rays, as illustrated in \figname\ref{fig:motivation}, there are large and small anatomical structures, such as the right/left lung, heart, and spinous processes; lung diseases can be local or global. This paper seeks to answer this critical question: {\it How to autodidactically learn generic source models from global and local patterns in health and disease?}

To answer this question, we have developed a novel SSL framework, called PEAC (patch embedding of anatomical consistency), to exploit global and local patterns in health and disease. As illustrated in \figname\ref{fig:cropping}-\ref{fig:architecture}, PEAC has an architecture of student-teacher, taking two global crops, one for the student and the other for the teacher, with overlaps from a chest X-ray to learn the global consistency between the two global crops and \jlsecnote{this is not what you implemented??}{the local consistency between each pair of corresponding local patches within the overlapped region of the two global crops}.\jlcleandelete{We have built PEAC on POPAR~\cite{pang2022popar} because all code of POPAR is publicly released, and it learns high-level anatomical knowledge via patch order prediction and gleans fine-grained features via (patch) appearance recovery. The global and local consistencies can be naturally implemented on top of POPAR.} Our extensive experiments have 
demonstrated\jlreplace{
    \jlnote{please summarize experiment results here}{that our PEAC gets SoTA which outperforms models pretrained on the fully-supervised dataset and previous self-supervised models. Also Our model consistently and robustly represents similar local anatomical structures at the patch level across diverse patients, while also providing a consistent representation of the overall anatomical structure.  }.
}{
    that our PEAC outperforms fully-supervised pretrained models on ImageNet or ChestX-ray14 and SoTA SSL methods, and offers 
    consistent representation of similar anatomical structures across diverse patients of different genders and weights and across different views of the same patient. 
}

Through this work, we have made the following contributions:
\begin{itemize}
    \item A straightforward yet effective SSL scheme that captures both global and local patterns embedded within medical images;
    \item A precise and stable patch-matching method that achieves anatomical embedding consistency in health and disease;
    \item Extensive illustrations that show the capability of PEAC in matching anatomical structures across different patients and across different views of the same patient and in segmenting anatomical structures by zero-shot;
    \item Thorough experiments that demonstrate the transferability of PEAC to various target tasks, outperforming SoTA full-supervised and self-supervised methods in classification and segmentation.
\end{itemize}

}
\jlnote{
{\color{red} \jlnote{check the new version}{Can we just mention that built PEAC on POPAR to learn high-level anatomical knowledge via patch order prediction and gleans
fine-grained features via (patch) appearance recovery. Because in one paragraph POPAR shows 4 times and reviewer may not know what POPAR is and downgrade our the specificity of our work.} \jlnote{We should {\bf keep the PEAC GitHub private} till the patent application is completed.}{Also about code stuff of POPAR we can mention in our github.}\\}
\jlnote{Please inject whatever I missed from your text into the (new) introduction}{{\color{red} 3 rows exceeds the 9 pages if do not delete something,}}
}{}

\section{Related Work}


\jlnote{The novelty of PEAC is very limited in its current shape. I suggest that you quickly move on to the second step: implementing the hierarchical consistency as I designed to demonstrate the ``locality'' and ``compositionality'' of PEAC}{}

\begin{figure}[h!]
    \centering
    \includegraphics[width=8cm]{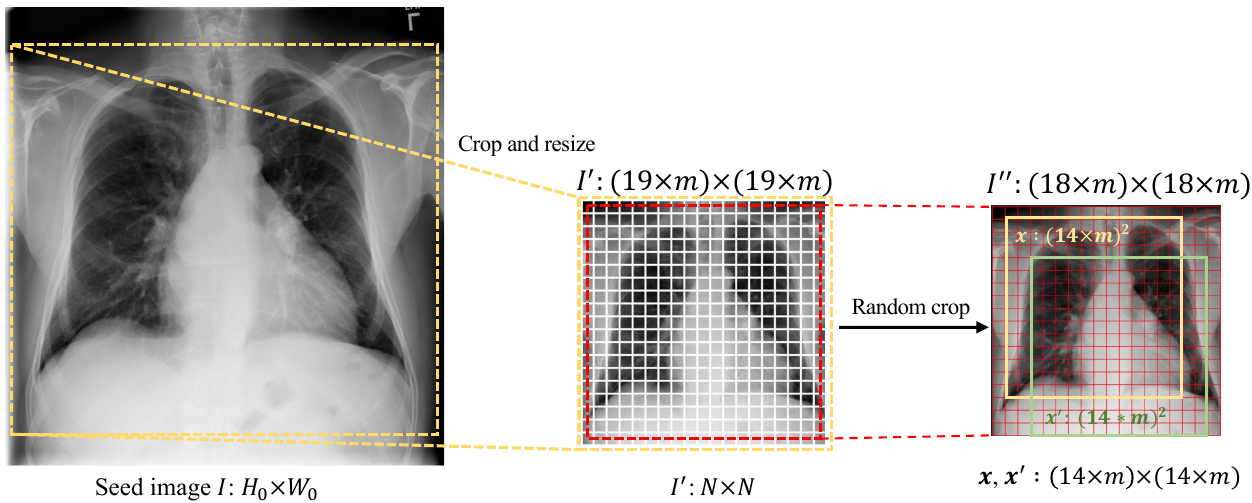}
    \caption{Grid-wise cropping\jlcleandelete{in image pre-processing} \jlthirdinsert{for stable grid-based matching}.
        \jlsecreplace{Firstly, a seed image is cropped from an original chest X-ray and resized to Image $I^{'}$ of size $(19 \times m) \times (19 \times m)$, so that $I^{'}$ can be conveniently partitioned to $19\times19$ patches with each patch of size $m \times m$. By default, $m=32$ in PEAC and PEAC$^{-1}$ with the Swin architecture, $m=16$ in PEAC$^{-2}$ with the Swin architecture and in PEAC$^{-3}$ with the ViT architecture. Secondly, Image $I^{''}$ with size $(18 \times m) \times (18 \times m)$ is randomly cropped from $I^{'}$ to ensure the diversity of local patches. Thirdly, Crops $x$ and $x^\prime$ of size $(14 \times m) \times (14 \times m)$ are randomly extracted from Image $I^{''}$ in alignment with the grid of Image $I^{''}$ to ensure the correspondence of local matches in the overlapped region between Crops $x$ and $x^\prime$. 
        \textcolor{blue}{Firstly, a seed image is cropped from an original chest X-ray and resized to Image $I^{'}$ of size $(n \times m) \times (n \times m)$, so that $I^{'}$ can be conveniently partitioned to $n\times n$ patches with each patch of size $m \times m$. By default, $n=19$, $m=32$ in PEAC and PEAC$^{-1}$ with the Swin architecture, $m=16$ in PEAC$^{-2}$ with the Swin architecture and in PEAC$^{-3}$ with the ViT architecture. Secondly, Image $I^{''}$ with size $( (n - 1) \times m) \times ( (n - 1) \times m)$ is randomly cropped from $I^{'}$ to ensure the diversity of local patches. Denotes input image size of the model as $s$. Thirdly, Crops $x$ and $x^\prime$ of size $((s \div m )\times m) \times ((s \div m ) \times m)$ are randomly extracted from Image $I^{''}$ in alignment with the grid of Image $I^{''}$ to ensure the correspondence of local matches in the overlapped region between Crops $x$ and $x^\prime$.}}
{Firstly, a seed image $I$ is cropped from an original chest X-ray and resized to Image $I'$ of size $(n \times m) \times (n \times m)$, so that $I'$ can be conveniently partitioned to $n\times n$ patches with each patch of size $m \times m$. By default, $n=19$ and $m=32$ in PEAC. Secondly, Image $I''$ with size $( (n - 1) \times m) \times ( (n - 1) \times m)$ is randomly cropped from $I'$ to ensure \jlthirdreplace{the}{a large} diversity of local patches \jlthirdinsert{during training}. Thirdly, Crops $x$ and $x^\prime$ of size $(k \times m) \times (k \times m)$ are randomly extracted from Image $I''$ in alignment with the grid of Image $I''$ to ensure the exact correspondence of local matches in the overlapped region between Crops $x$ and $x^\prime$ (referred to stable grid-based matching and detailed in~\secname~\ref{sec:ablations}). By default, $k=14$ in PEAC. See the Appendix for various PEAC configurations.}
    }
    \label{fig:cropping}
    \vspace{-0.2cm}
\end{figure}

\begin{figure}[h!]
    \centering
    \vspace{-0.1cm}
    \setlength{\belowcaptionskip}{0.1cm}
    \includegraphics[width=12cm]{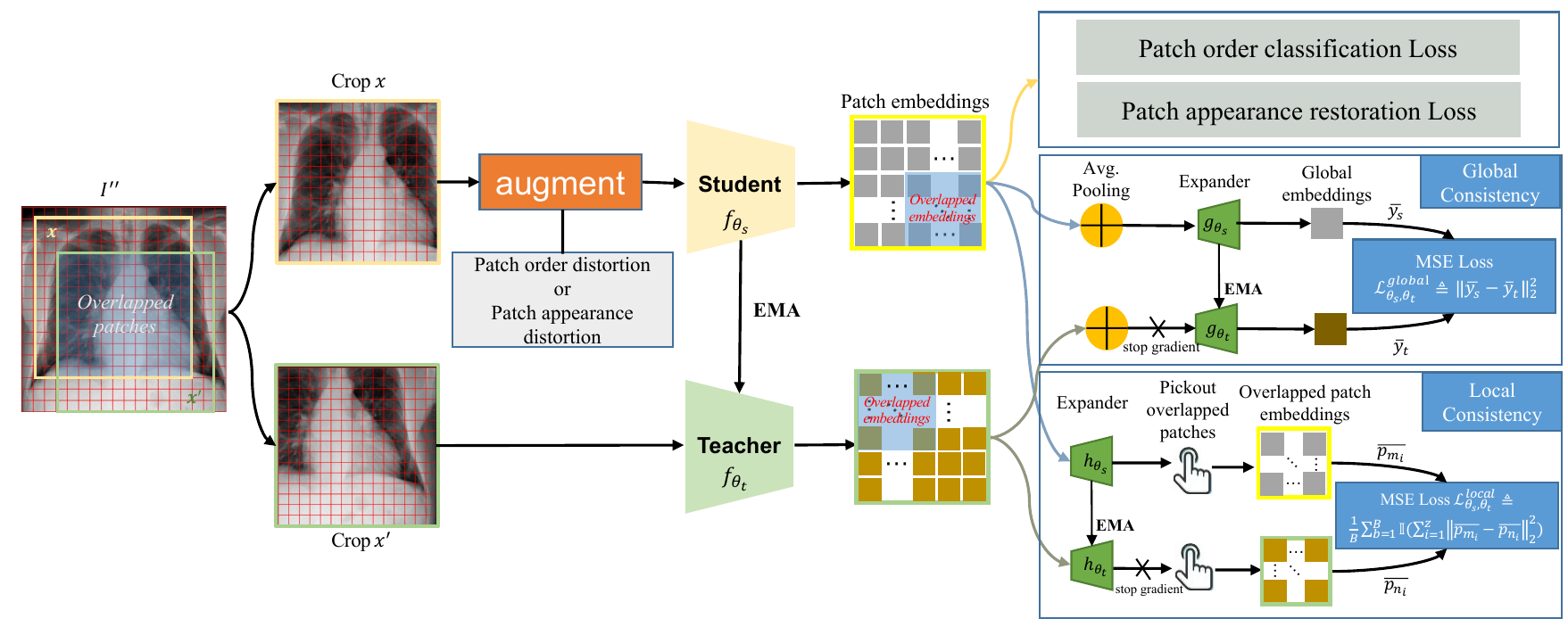}
    \caption{\jlsecnote{see my {\bf new} comments sent to you via email}{}\jlsecreplace{An overview of our proposed method. Two cropped images are fed into the student-teacher network. And we combine two modules: The student branch stands along learns high-level anatomical structures and fine-grained image features by patch order classification loss and patch appearance restoration loss, respectively; Combining the teacher branch and the student branch to learn consistent contextualized high-level anatomical structures and fine-grained local anatomical structures across different views.}
{PEAC has an architecture of student-teacher, taking two global crops: $x$ for the student and $x^\prime$ for the teacher, with overlaps from a chest X-ray to learn the global consistency (\eqname\ref{eq5}) between the two global crops ($x$ and $x^\prime$) and \jlsecnote{not as you implemented?? same with our implementation}{the local consistency (\eqname\ref{eq7}) between each pair of corresponding local patches within the overlapped region of the two global crops ($x$ and $x^\prime$)}. The student, built on POPAR~\cite{pang2022popar}, learns high-level relationships among anatomical structures by patch order classification and fine-grained image features by patch appearance restoration, as detailed in the Appendix. Integrating the teacher with the student aims to learn consistent contextualized embedding for coarse-grained global anatomical structures and fine-grained local anatomical structures across different views of the same patients, leading to anatomically consistent embedding across patients.
    }
    }
    \label{fig3} 
    \label{fig:architecture}
    \vspace{-0.3cm}
\end{figure}

{\bf Global features and \jlsecnoterem{no reason to capitalize}{local} features.} Global features describe the overall appearance of the image. Most recent methods for global feature learning are put forward to ensure that the extracted global features are consistent across different views. The methods to achieve this include contrastive learning and non-contrastive learning methods. Contrastive methods~\cite{he2020momentum, oord2018representation, tian2020contrastive, henaff2020data, hjelm2018learning, li2020prototypical} bring representation of different views of the same image closer and spreading representations of views from different images apart. Non-contrastive methods rely on maintaining the informational content consistent of the representations by either explicit regularization~\cite{zbontar2021barlow, bardes2022variance, bardes2022vicregl} or architecture design like Siamese architecture~\cite{chen2021exploring, grill2020bootstrap, lee2021compressive}. In opposition to global methods, local features describe the information that is specific to smaller regions of the image. In local features learning methods, a contrastive or consistent loss can be applied directly at the pixel level~\cite{xie2021propagate}, the feature map level~\cite{wang2021dense, bardes2022vicregl} or the image region level~\cite{xiao2021region} which forces consistency between pixels at similar locations, between groups of pixels and between large regions that overlap in different views of an image. However, at present the vast majority of methods that use local features calculate embedding consistency or contrastive learning loss based on the relative positions of the features~\cite{bardes2022vicregl, yun2022patch, yi2020patch}, such as the feature vectors of semantically closest patches or spatially nearest neighbor patches. In contrast, our PEAC method calculates the consistency loss based on the absolute positions of overlapping image patches shown in \figname \ref{fig:cropping}. In this way, fine-grained anatomical structures can be more accurately characterized. 


\section{Our Method}


\jlthirdnote{In \figname\ref{fig:architecture}, please indicate the components with their corresponding math symbols used in the text}{}

\jlthirdnote{Have you worked with the writing center on this paragraph?}{The goal of our method, Patch Embedding of Anatomical Consistency (PEAC), is to learn global and local anatomical structures underneath medical images. In medical images, there are various local patterns such as spinous processes, clavicle, mainstem bronchus, hemidiaphragm, the osseous structures of the thorax, etc. The \jlthirdnote{what is ``analogous''?}{resemblances} can be captured by the two global crops shown in \figname\ref{fig3} so that global embedding consistency can encourage the network to extract high-level semantic features of similar local regions. Besides, local embedding consistency based grid-like image patches can equip the model with a more stable matching strategy, for disease diagnosing which needs both single and multiple local patterns to catch the fine-grained anatomical structure.
Therefore, we proposed a network that considers both global and local features of medical images at the same time.
}

As shown in \figname\ref{fig3}, PEAC is \jlsecreplace{a}{an} SSL framework comprised of four key components: (1) Student-Teacher model that aims to extract features of two crops simultaneously; \jlthirdnote{simplify and straighten the logic}{(2) image augmentation and restoration module aim to restore image crops from the \jlthirdnoterem{should they be order distortion and appearance distortion? Use consistent terminologies throughout the paper; define them in their first use and use them consistently}{patch order and appearance distortion};} (3) global module that aims to enforce the model to learn coarse-grained global features of two crops; (4) local module that aims to enforce the model to learn fine-grained local features from overlapped patches. By integrating the above modules, the model learns the coarse-grained, fine-grained and contextualized high-level anatomical structure features. In the following, we will introduce our methods from image pre-processing, each components and the joint training loss. Our model is based on POPAR because we need to model the overall structural information and local detailed and robust information of medical images.

\subsection{Global Embedding Consistency}

\jlthirdnote{We should consider evaluating different losses for global and local consistencies as ablation studies, such as those in DINO, MoCo, and Barlow Twins. Actually, \ul{the loss used in Barlow Twins could be the most effective one for implementing PEAC}??}{}
Before inputting to the model, the seed images are pre-processed in grid-wise cropping to get two crops $x$, $x^\prime \in R^{C \times H \times W}$, $C$ is the number of channels, $(H, W)$ are the crops' spatial dimensions, shown in \figname\ref{fig:cropping}. Then the two crops are input to the \emph{Student} and \emph{Teacher} encoders $f_{\theta_s}, f_{\theta_t}$ to get the local features $s$, $t$ respectively. Then in the global branch the average pooling operators $\oplus: {R}^{D \times H \times W} \rightarrow {R}^D$ are performed on the local features. We denote the pooled representations as $y_{s\oplus}$ and $y_{t\oplus} \in R^D$. At last the expanders $g_{\theta_{s}}$, $g_{\theta_{t}}$ are 3-layer MLP which map $y_{s\oplus}$, $y_{t\oplus}$ to get the embedding vectors $y_s, y_t \in R^{H}$. We put the $l_2$-normalize to $\overline{y_s}=y_s/\| y_s\|_2$ and $\overline{y_t}=y_t/ \|y_t \|_2$. \jlthirdnote{put $y_s$, $\overline{y_s}$, $y_t$, $\overline{y_t}$ in Figure 3 and ``(Equation 1)'' next to ``MSR Loss''}{}. At last, we define global patch embedding consistency loss as the following mean square error between the normalized output,
\begin{equation} \label{eq5} 
    \mathcal{L}^{global}_{\theta_s, \theta_t} \triangleq \|\overline{y_s} - \overline{y_t}\|_2^2 = 2-2\cdot \frac{\left\langle y_s,y_t \right\rangle}{\| y_s\|_2 \cdot \|y_t \|_2}
\end{equation}

We symmetrize the loss from \eqname\ref{eq5} by separately feeding $x$ to \emph{Teacher} encoder and $x^\prime$ to \emph{Student} encoder to compute $\widetilde{\mathcal{L}}^{global}_{\theta_s, \theta_t}$. Accordingly, we get the global loss as $\mathcal{L}^{G}_{\theta_s, \theta_t} = \mathcal{L}^{global}_{\theta_s, \theta_t}+\widetilde{\mathcal{L}}^{global}_{\theta_s, \theta_t}$.

\subsection{Local Embedding Consistency}
\jlthirdnote{Please consider an illustrate to make this section on ``Local Embedding Consistency'' easier to understand}{}
As the encoders are Vision Transformer network, the crop is divided into a sequence of $N$ non-overlapping image patches $P = (p_1, p_2, ...,p_N)$ where $N = \frac{H \times W}{m^2}$ and $m$ is the patch resolution. The encoder of the Student-Teacher model extracts local features $ s,t \in {R}^{D \times N}$ from the two crops $x, x^\prime$. We denote $s_{k}$ and $t_{k} \in R^D$ the feature vectors at position $k \in [1, ..., N]$ in their corresponding feature maps. Since the image patches are randomly sampled from an image grid with an overlap rate of 50\%-100\%, we define the overlapping image patches $O_{m}$, $O_{n}$ for $x$ and $x^\prime$ respectively, and $m \in [m_1, ..., m_z]$, $n \in [n_1, ...,n_z]$ are the patch indexes of the overlapping region, $z$ is the number of overlapping patches. $O_{m_i}$ and $O_{n_i}$ are in correspondece where $1 \leq i \leq z$ and we call this process grid matching. Correspondingly, $O_{m}$ and $O_{n}$ are transformed into embedding vectors $o_{m}$ and $o_{n}$ through the feature extractors. Then in the local module there are 3-layer MLP expanders $h_{\theta_{s}}$, $h_{\theta_{t}}$ adding to $o_{m}$, $o_{n}$ to get the final local patch embedding vectors $p_{m}$, $p_{n}$. Similarly, we also put $l_2$-normalize to $\overline{p_{m}}=p_{m}/\| p_{m}\|_2$, $\overline{p_{n}}=p_{n}/\| p_{n}\|_2$. We only randomly add patch order distortion and patch appearance distortion in the student branch. When the patch order is distorted, the patch embedding vector will represent the distorted global feature for attention mechanism. And local embeddings of distorted and non-distorted patch orders in the student and teacher branches can't be consistent. So we won't compute local loss if the crop gains patch order distortion (indicator $\mathbb{I}=0$) while it has no impact on the patch appearance distortion ($\mathbb{I}=1$). To align the output of the student and teacher networks regarding local features, we define the following local patch embedding consistency loss function in \eqname\ref{eq7}
\begin{equation} \label{eq7} 
    \mathcal{L}^{local}_{\theta_s, \theta_t} \triangleq \frac{1}{B} \sum_{b=1}^B \mathbb{I} \cdot (\sum_{i=1}^z \|\overline{p_{m_i}} - \overline{p_{n_i}}\|_2^2)
\end{equation}
$\overline{p_{m_i}}$ and $\overline{p_{n_i}}$ are the embedding vectors of the $i$-th overlapping image patches and $B$ is the batch size. Similar to the global loss in previous section, when $x$ is fed into \emph{Teacher} encoder and $x^\prime$ is fed into \emph{Student} encoder, we compute corresponding loss $\widetilde{\mathcal{L}}^{local}_{\theta_s, \theta_t}$. So the local loss $\mathcal{L}^{L}_{\theta_s, \theta_t} = \mathcal{L}^{local}_{\theta_s, \theta_t}+\widetilde{\mathcal{L}}^{local}_{\theta_s, \theta_t}$. 
\jlthirdnote{put $p_{m_i}$, $p_{n_i}$, $\overline{p_{m_i}}$, and $\overline{p_{n_i}}$ in Figure 3 and ``(Equation 2)'' next to ``MSR Loss''}{}
\jlthirdnote{The locality property of PEAC seems weaker that that of Adam/Eve. Please consider a loss for the dissimilarity between the nonoverlapped regions, that is, we pull the overlapped regions together and push the nonoverlapped regions apart.}{}

We calculate $\mathcal{L}^{oc}_{\theta_s}=-\frac{1}{B} \sum_{b=1}^B \sum_{l=1}^n \sum_{c=1}^n \mathcal{Y} \log \mathcal{P}^{o}$ and $\mathcal{L}^{ar}_{\theta_s}=\frac{1}{B} \sum_{i=1}^B \sum_{j=1}^n\left\|p_j-p_j^{a}\right\|_2^2$ for patch order distortion and patch appearance distortion in the student branch. Where $n$ is the number of patches for each image, $\mathcal{Y}$ represent the order ground truth and $\mathcal{P}^{o}$ represent the network's patch order prediction, $p_j$ and $p_j^{a}$ represent image original appearance and reconstruction prediction.


Finally, the total loss is defined in \eqname\ref{eq: total loss}, where $\mathcal{L}^{oc}_{\theta_s}$ is patch order classification loss, $\mathcal{L}^{ar}_{\theta_s}$ is patch appearance restoration loss, $\mathcal{L}^{G}_{\theta_s, \theta_t}$ is the global patch embedding consistency loss and $\mathcal{L}^{L}_{\theta_s, \theta_t}$ is the local patch embedding consistency loss. $\mathcal{L}^{oc}_{\theta_s}$ and $\mathcal{L}^{ar}_{\theta_s}$ empower the model to learn high-level anatomical structures. The $\mathcal{L}^{G}_{\theta_s, \theta_t}$ equips the model to learn the coarse-grained granularity and synthetical anatomy from global patch embeddings. $\mathcal{L}^{L}_{\theta_s, \theta_t}$ lets the model learn fine-grained and precise anatomical structures from local patch embeddings of overlapped parts. $\overline{y_s}$



\begin{equation} \label{eq: total loss} 
\jlthirdnote{\text{Consider cyclic pretraining instead of summing up all losses as one loss}}{}
   \mathcal{L}=\mathcal{L}^{oc}_{\theta_s}+\mathcal{L}^{ar}_{\theta_s}+\mathcal{L}^{G}_{\theta_s, \theta_t}+\mathcal{L}^{L}_{\theta_s, \theta_t}
\end{equation}

\section{Experiments}

\jlsecnote{I did not see the performance of PEAC (the final product) but downgraded versions of PEAC??} 

\subsection{Datasets and Implementation Details}

\textbf{Pretraining Settings.} We pretrain PEAC with Swin-B as the backbone on unlabeled \jlthirdnote{Please work with Jiaxuan closely to start large-scale pretraining with >800K chest X-rays organized by him and DongAo. }{} ChestX-ray14~\cite{wang2017chestx} dataset. 
Our PEAC and PEAC$^{-1}$ models utilize Swin-B as the backbone, pre-trained on an image size of $448 \times 448$ and fine-tuned on $448 \times 448$ and $224 \times 224$ respectively. PEAC$^{-3}$ adopts ViT-B as the backbone, pre-trained and fine-tuned on an image size of $224 \times 224$. As for the prediction heads in the student branch, we use two single linear layers for the classification (patch order) and restoration tasks (patch appearance), and two 3-layer MLP for the expanders of local and global features. The augments used in the student branch include 50\% probability of patch appearance distortion~\cite{zhou2021models} and 50\% probability of shuffling patches. The weights of Student model are updated by the total loss while the Teacher model are updated by Exponential Moving Average (EMA)~\cite{tarvainen2017mean} after each iteration. More details are in the Appendix.

\smallskip
\noindent \textbf{Target Tasks and Datasets.} To assess the performance of PEAC models, we transfer them to four thoracic disease classification tasks ChestX-ray14~\cite{wang2017chestx}, CheXpert~\cite{irvin2019chexpert}, NIH Shenzhen CXR~\cite{jaeger2014two}, RSNA Pneumonia~\cite{rsna} and one chest organ segmentation task JSRT~\cite{shiraishi2000development}. For the segmentation task, we integrate Upernet~\cite{xiao2018unified} into the process. All downstream models share the same Swin-B backbone, where the encoder is initialized with PEAC pretrained weights and a final prediction decoder is re-initialized based on the number of classes for the target task. The total training rounds for classification and segmentation are 150 and 500 epochs, respectively. More details can be found in Appendix

\subsection{Experimental Results}

\jlnoterem{use spaces properly.}{\textbf{(1) PEAC outperforms fully-supervised pretrained models. }We} compare the performance of PEAC on four downstream tasks with different initialization methods.\jlsecnoterem{wordy and logic back and forth}{} As shown in \tablename\ref{tab: fully-supervised}, PEAC model outperforms both supervised ImageNet and ChestX-ray14 models, demonstrating that PEAC has learned transferable features for various medical image tasks. To show the statistical significance, we have conducted t-tests between PEAC/PEAC$^{-1}$ and ImageNet-21K on all target tasks, yielding p-values: (\jlnoterem{please include p-values here}{$3.2\times10^{-9}, 4.0\times10^{-10}, 5.9\times10^{-5}, 0.018$})/($2.9\times10^{-4}, 4.2\times10^{-8}, 0.033, 0.027$) (all p-values$<$0.05), showing that the performance gains by PEAC, even by PEAC$^{-1}$, on all four target tasks are statistically significant at level of 0.05.

\jlsecreplace{
    \begin{table}[hbp]
        \centering
        \setlength{\belowcaptionskip}{0.1cm}
        \caption{\jlnote{downgraded versions need to be defined before using them. The methods listed under ``Initialization'' do not belong to the same category; consider making them conisstent}{} \jlsecnote{see my comments sent via email}{} PEAC models outperform fully-supervised pretrained models on ImageNet and ChestX-ray14 datasets in three target tasks cross architectures. The best methods are bolded while the second best are underlined. Transfer learning is inapplicable when pretraining and target tasks are the same, denoted by "-"}
        \label{tab: fully-supervised} 
        \resizebox{.8\columnwidth}{!}{
        \begin{tabular}{cccccc}
        \toprule    
             \multirow{2}*{Backbone} & Pretraining data & \multirow{2}*{ChestX-ray14} & \multirow{2}*{CheXpert} & \multirow{2}*{ShenZhen} & RSNA  \\
              & and methods& & & & Pneumonia \\
        \midrule
            \multirow{3}*{ResNet-50} & Random & 80.40 $\pm$ 0.05 & 86.60 $\pm$ 0.17 & 90.49 $\pm$ 1.16 & 70.00 $\pm$ 0.50 \\
            & ImageNet-1K & 81.70 $\pm$ 0.15 & 87.17 $\pm$ 0.22 & 94.96 $\pm$ 1.19 & 73.04 $\pm$ 0.35\\
            & ChestX-ray14 & - & 87.40 $\pm$ 0.26 & 96.32 $\pm$ 0.65 & 71.64 $\pm$ 0.37\\
        \midrule
            \multirow{4}*{ViT-B} & Random & 70.84 $\pm$ 0.19 & 80.78 $\pm$ 0.13 & 84.46 $\pm$ 1.65 & 66.59 $\pm$ 0.39\\
            & ImageNet-21K & 77.55 $\pm$ 1.82 & $83.32 \pm 0.69$ & 91.85 $\pm$ 3.40 & 71.50 $\pm$ 0.52 \\
            &ChestX-ray14 & - & $84.37 \pm 0.42$ & $91.23 \pm 0.81$ & $66.96 \pm 0.24$ \\
            & PEAC$^{-3}$ & $80.04 \pm 0.20$ & 88.10\pm 0.29 & $96.69 \pm 0.30$ & $\underline{73.77 \pm 0.39}$\\
        \midrule
             \multirow{5}*{Swin-B} & Random & 74.29 $\pm$ 0.41 & 85.78 $\pm$ 0.01 & 85.83 $\pm$ 3.68 & 70.02 $\pm$ 0.42 \\
             & ImageNet-21K & 81.32 $\pm$ 0.19 & $87.94 \pm 0.36$ & 94.23 $\pm$ 0.81 & 73.15 $\pm$ 0.61 \\
             &ChestX-ray14 & - & $87.22 \pm 0.22$ & $91.35 \pm 0.93$ & $70.67 \pm 0.18$ \\
             & PEAC$^{-1}$ & $81.90 \pm 0.15$ & $88.64 \pm 0.19$& $97.17 \pm 0.42$ & $73.70 \pm 0.48$\\
             & PEAC & $\bm{82.78} \pm \bm{0.21}$ & $\bm{88.81} \pm \bm{0.57} $& $\bm{97.39} \pm \bm{0.19}$ & $\bm{74.39 \pm 0.66}$\\
    
        \bottomrule
        \end{tabular}}
    \end{table}
}{
    \begin{table}[hbp]
        \centering
        \setlength{\belowcaptionskip}{0.1cm}
        \caption{PEAC models outperform fully-supervised pretrained models on ImageNet and ChestX-ray14 datasets in four target tasks across architectures. The best methods are bolded while the second best are underlined. We have conducted independent two sample \textit{t}-text between the best vs. others and highlighted \jlthirdreplace{boxes}{those} in blue when they are not significantly different at $p=0.05$ level. Transfer learning is inapplicable, when pretraining and target tasks are the same, and denoted by "--". \jlthirdnote{compute p-values and highlight rows as in \figname~\ref{seg results}.}{}}
        \label{tab: fully-supervised} 
        \resizebox{.95\columnwidth}{!}{
        \begin{tabular}{ccccccc}
        \toprule    
             Backbone & Pretraining data & Pretraining method & ChestX-ray14 & CheXpert & ShenZhen & RSNA Pneumonia \\
        \midrule
            \multirow{3}*{ResNet-50} & \multicolumn{2}{c}{No pretraining (i.e., training from scratch)} & 80.40 $\pm$ 0.05 & 86.60 $\pm$ 0.17 & 90.49 $\pm$ 1.16 & 70.00 $\pm$ 0.50 \\
            & ImageNet-1K & Fully-supervised & 81.70 $\pm$ 0.15 & 87.17 $\pm$ 0.22 & 94.96 $\pm$ 1.19 & 73.04 $\pm$ 0.35\\
            & ChestX-ray14 & Fully-supervised & -- & 87.40 $\pm$ 0.26 & 96.32 $\pm$ 0.65 & 71.64 $\pm$ 0.37\\
        \midrule
            \multirow{4}*{ViT-B} & \multicolumn{2}{c}{No pretraining (i.e., training from scratch)} & 70.84 $\pm$ 0.19 & 80.78 $\pm$ 0.13 & 84.46 $\pm$ 1.65 & 66.59 $\pm$ 0.39\\
            & ImageNet-21K & Fully-supervised  & 77.55 $\pm$ 1.82 & $83.32 \pm 0.69$ & 91.85 $\pm$ 3.40 & 71.50 $\pm$ 0.52 \\
            &ChestX-ray14 & Fully-supervised  & -- & $84.37 \pm 0.42$ & $91.23 \pm 0.81$ & $66.96 \pm 0.24$ \\
            & ChestX-ray14 & PEAC$^{-3}$ (self-supervised)  & $80.04 \pm 0.20$ & 88.10 $\pm$ 0.29 & $96.69 \pm 0.30$ & $\underline{73.77 \pm 0.39}$\\
        \midrule
             \multirow{5}*{Swin-B} & \multicolumn{2}{c}{No pretraining (i.e., training from scratch)} & 74.29 $\pm$ 0.41 & 85.78 $\pm$ 0.01 & 85.83 $\pm$ 3.68 & 70.02 $\pm$ 0.42 \\
             & ImageNet-21K & Fully-supervised & 81.32 $\pm$ 0.19 & $87.94 \pm 0.36$ & 94.23 $\pm$ 0.81 & 73.15 $\pm$ 0.61 \\
             &ChestX-ray14 & Fully-supervised & -- & $87.22 \pm 0.22$ & $91.35 \pm 0.93$ & $70.67 \pm 0.18$ \\
             & ChestX-ray14 & PEAC$^{-1}$ (self-supervised) & $\underline{81.90 \pm 0.15}$ & \cellcolor{cyan!25}$\underline{88.64 \pm 0.19}$& $\underline{97.17 \pm 0.42}$ & $73.70 \pm 0.48$\\
             & ChestX-ray14 & PEAC (self-supervised) & $\bm{82.78} \pm \bm{0.21}$ & \cellcolor{cyan!25}$\bm{88.81} \pm \bm{0.57} $& $\bm{97.39} \pm \bm{0.19}$ & $\bm{74.39 \pm 0.66}$\\
    
        \bottomrule
        \end{tabular}}
    \end{table}
}

\begin{wraptable}{r}{5.3cm}
    \centering
    \vspace{-1.0em}
    \setlength{\abovecaptionskip}{0cm}
    \setlength{\belowcaptionskip}{0.1cm}
    \caption{Comparing PEAC with SoTA baselines in terms of sensitivity to the number of training samples on ChestX-ray14. \jlthirdreplace{}{The baseline performance is adopted from DiRA~\cite{haghighi2022dira}.}}
    \label{limited_annotation}
    \resizebox{0.38\columnwidth}{!}{
    \begin{tabular}{cccc}
    \toprule    
          Method & $25\%$ & $50\%$  & $100\%$ \\
    \midrule
        MoCo-v2~\cite{chen2020improved, haghighi2022dira}   &  74.71  & 76.89  & 80.36 \\
        Barlow Twins~\cite{zbontar2021barlow, haghighi2022dira} & 76.23   & 77.59& 80.45\\
        SimSiam~\cite{chen2021exploring, haghighi2022dira} & 73.05   & 75.20  & 79.62\\
        DiRA$_{\text{MoCo-v2}}$~\cite{haghighi2022dira} & $\underline{77.55}$   & $\underline{78.74}$  & $\underline{81.12}$ \\
        PEAC & \textbf{77.78}  & \textbf{79.29}   & \textbf{82.78} \\
    \bottomrule
    \end{tabular}
    }
    \vspace{-0.15cm} 
\end{wraptable}

\smallskip
\noindent
\textbf{(2) PEAC outperforms self-supervised models pretrained on ImageNet.} 
    We compared PEAC models pretrained on ChestX-rays14 with the transformer-based models pretrained on ImageNet by various SOTA self-supervised methods including MoCo V3~\cite{chen2021empirical}, SimMIM~\cite{xie2022simmim}, DINO~\cite{caron2021emerging}, BEiT~\cite{bao2021beit}, and MAE~\cite{he2022masked}. The results in \tablename\ref{tab:imagenet} show that
    our PEAC pretrained on moderately-sized unlabeled chestX-rays14 yields better results than the aforementioned SSL methods pretrained on larger ImageNet, \jlsecreplace{which \jlsecnote{Experiments cannot approve things}{proves} the effectiveness of pretraining PEAC with in-domain medical data.
}{
revealing the transferability of PEAC via learning anatomical consistency with in-domain medical data.
    
}    

\smallskip
\noindent
\textbf{(3) PEAC outperforms recent self-supervised models \jlinsert{pretrained on medical images.}} To demonstrate the effectiveness of representation learning via our proposed framework, we compare PEAC with SoTA CNN-based (SimSiam~\cite{chen2021exploring}, MoCo V2~\cite{chen2020improved}, Barlow Twins~\cite{zbontar2021barlow}) and transformer-based SSL methods (SimMIM~\cite{xie2022simmim}) pretrained on medical images as shown in \tablename\ref{tab: others}. We also compare PEAC with these SoTA baselines in small data regimes shown in \tablename\ref{limited_annotation}. The downstream dataset ChestX-ray14 is randomly seleted 25\% and 50\% when fintuning. \jlsecnote{wordy and logic back and forth}{Our method yields the best performance across four datasets and verifies its effectiveness on different backbones. We get several observations from the results: (1) In transformer backbones, our methods outperform SimMIM which demonstrates that patch order distortion and patch embedding consistency are effective; (2) In Swin-B backbone, our method outperforms POPAR$^{-1}$ which shows the good generalization performance of patch embedding consistency; (3) For different downstream label fractions, our method outperforming other baselines shows the robustness of patch embedding consistency.}

\begin{wraptable}{r}{8cm}
    \centering
    \vspace{-0.3cm}
    \setlength{\abovecaptionskip}{0cm}
    \setlength{\belowcaptionskip}{0.1cm}
    \caption{Our PEAC \jlthirdreplace{lead}{leads} the best or comparable performance on segmentation dataset \jlreplace{JART}{JSRT} over a self-supervised learning approach POPAR and fully supervised pretrained model. }
    \label{seg results}
    \resizebox{0.6\columnwidth}{!}{
    \begin{tabular}{ccccc}
    \toprule
         Pretrain Method & Pretrain Dataset & JSRT Lung & JSRT Heart & JSRT Clavicle\\
    \midrule
        Scratch & None& 97.48 $\pm$ 0.08 & 92.87 $\pm$ 1.31& 87.52 $\pm$ 1.78\\
        Supervised&ImageNet-1K & $\cellcolor{cyan!25}\bm{97.99\pm0.01}$& \cellcolor{cyan!25}95.12 $\pm$ 0.08 & \cellcolor{cyan!25}92.18 $\pm$ 0.02\\
        POPAR&ChestX-ray14 & \cellcolor{cyan!25}97.88 $\pm$ 0.12 & 93.67 $\pm$ 0.95& 91.78 $\pm$ 0.53\\
        PEAC& ChestX-ray14& \cellcolor{cyan!25}$97.97 \pm 0.01$ & \cellcolor{cyan!25}$\bm{95.21\pm0.02}$& \cellcolor{cyan!25}$\bm{92.19\pm0.04}$\\
    \bottomrule
    \end{tabular}
    }
    \vspace{-0.20cm}
\end{wraptable}

\smallskip
\noindent
\textbf{(4) PEAC exhibits prominent transferability for segmentation tasks.} \jlthirdnote{Wordy -- simplify -- have you worked with the writing center on this paragraph? No, it's newly updated}{As shown in \tablename\ref{seg results}, PEAC (1) surpasses SoTA SSL method POPAR~\cite{pang2022popar} on two organ segmentation tasks (2) pretrained on the moderate-size Chest X-ray dataset yields competitive performance akin to fully supervised ImageNet models. These outcomes emphasize that employed as the backbone weights in the segmentation network, PEAC models facilitate accurate pixel-level predictions across three organ segmentation tasks. More details are shown in \figname\ref{seg_pvalue} of Appendix.}


\begin{table}[hbp]
    \centering
    \setlength{\belowcaptionskip}{0.2cm}
    \caption{Even downgraded PEAC$^{-1}$ and PEAC$^{-3}$\jlcleandelete{(two downgraded versions of PEAC)} \jlreplace{outperforms}{outperform} SoTA self-supervised ImageNet in \jlsecreplace{three}{four} target tasks. The best results are bolded and the second best are underlined. \jlthirdnote{compute p-values and highlight rows as in \figname~\ref{seg results}.}{}}
    \label{tab:imagenet} 
    \resizebox{.9\columnwidth}{!}{
    \begin{tabular}{c|c|c|cccc}
    \toprule
         \multirow{2}*{Backbone} &Pretrained& \multirow{2}*{Method} & \multirow{2}*{ChestX-ray14} & \multirow{2}*{CheXpert} & \multirow{2}*{ShenZhen} & RSNA  \\
          &dataset& & & & & Pneumonia \\
    \midrule
        \multirow{6}*{ViT-B} &\multirow{5}*{ImageNet}& MoCo V3 & 79.20 $\pm$ 0.29 & 86.91 $\pm$ 0.77 & 85.71 $\pm$ 1.41 & 72.79 $\pm$ 0.52 \\
        && SimMIM & 79.55 $\pm$ 0.56 & 87.83 $\pm$ 0.46 & 92.74 $\pm$ 0.92 & 72.08 $\pm$ 0.47\\
        & &DINO & 78.37 $\pm$ 0.47 & 86.91 $\pm$ 0.44 & 87.83 $\pm$ 7.20 & 71.27 $\pm$ 0.45\\
        & &BEiT & 74.69 $\pm$ 0.29 & 85.81 $\pm$ 1.00 & 92.95 $\pm$ 1.25 & 72.78 $\pm$ 0.37\\
        & &MAE & 78.97 $\pm$ 0.65 & 87.12 $\pm$ 0.54 & 93.58 $\pm$ 1.18 & 72.85 $\pm$ 0.50\\
        \cline{2-7}
        & ChestX-ray14&PEAC$^{-3}$ & $80.04 \pm 0.20$ & 88.10 $\pm$ 0.29 & $96.69 \pm 0.30$ & $\underline{73.77 \pm 0.39}$\\
    \cline{1-7}
         \multirow{3}*{Swin-B} &ImageNet& SimMIM & $81.39 \pm 0.18$ & 87.50 $\pm$ 0.23 & 87.86 $\pm$ 4.92 & 73.15 $\pm$ 0.73 \\
         \cline{2-7}
         &\multirow{2}*{ChestX-ray14}& PEAC$^{-1}$ & $\underline{81.90 \pm 0.15}$ & \cellcolor{cyan!25}$\underline{88.64 \pm 0.19}$& $\underline{97.17 \pm 0.42}$ & $73.70 \pm 0.48$\\
         &&PEAC & $\bm{82.78} \pm \bm{0.21}$ & \cellcolor{cyan!25}$\bm{88.81} \pm \bm{0.57} $& $\bm{97.39} \pm \bm{0.19}$ & $\bm{74.39 \pm 0.66}$\\
    \bottomrule
    \end{tabular}}
\end{table}

\begin{table}[hbp]
    \centering
    \vspace{-0.2cm}
    \setlength{\belowcaptionskip}{0.2cm}
    \caption{
    To speed up the training process, we compare the performance of downstream tasks using image resolution of 224. 
    All models are pretrained on the ChestX-ray14 dataset. \jlthirdnote{compute p-values and highlight rows as in \figname~\ref{seg results}.}{}}
    \label{tab: others} 
    \resizebox{.9\columnwidth}{!}{
    \begin{tabular}{c|c|c|cccc}
    \toprule
         \multirow{2}*{Backbone} &Pretrained& \multirow{2}*{Method} & \multirow{2}*{ChestX-ray14} & \multirow{2}*{CheXpert} & \multirow{2}*{ShenZhen} & RSNA  \\
          & dataset&& & & & Pneumonia \\
    \midrule
        \multirow{3}*{ResNet-50} &\multirow{3}*{ChestX-ray14}& SimSiam & 79.62 $\pm$ 0.34 & 83.82 $\pm$ 0.94 & 93.13 $\pm$ 1.36 & 71.20 $\pm$ 0.60 \\
        && MoCoV2 & 80.36 $\pm$ 0.26 & 86.42 $\pm$ 0.42 & 92.59 $\pm$ 1.79 & 71.98 $\pm$ 0.82\\
        && Barlow Twins & 80.45 $\pm$ 0.29 & 86.90 $\pm$ 0.62 & 92.17 $\pm$ 1.54 & 71.45 $\pm$ 0.82\\
    \midrule
        \multirow{2}*{ViT-B} & \multirow{2}*{ChestX-ray14} & SimMIM & $79.20 \pm 0.19$ & $83.48 \pm 2.43$ & $93.77 \pm 1.01$ & $71.66 \pm 0.75$ \\
        &&PEAC$^{-3}$ & $80.04 \pm 0.20$ & 88.10 $\pm$ 0.29 & $96.69 \pm 0.30$ & $\cellcolor{cyan!25}\bm{73.77 \pm 0.39}$\\
    
    \midrule
         \multirow{3}*{Swin-B} &\multirow{3}*{ChestX-ray14}& SimMIM & 79.09 $\pm$ 0.57 & 86.75 $\pm$ 0.96 & 93.03 $\pm$ 0.48 & 71.99 $\pm$ 0.55 \\
         && $\rm POPAR^{-1}$ & $\underline{80.51 \pm 0.15}$ & $\underline{88.16 \pm 0.66}$ & $\underline{96.81 \pm 0.40}$ & 73.58 $\pm$ 0.18 \\
         
         && PEAC$^{-1}$ & $\bm{81.90 \pm 0.15}$ & $\bm{88.64 \pm 0.19}$& $\bm{97.17 \pm 0.42}$ & \cellcolor{cyan!25}$\underline{73.70 \pm 0.48}$\\
    \bottomrule
    
    \end{tabular}
    }
    \vspace{-0.20cm}
\end{table}

\subsection{Ablations}
\label{sec:ablations}

\begin{wrapfigure}{r}{6.7cm}
    \setlength{\belowcaptionskip}{0.1cm}
    \vspace{-0.1cm}
    \centering
    \includegraphics[width=7cm]{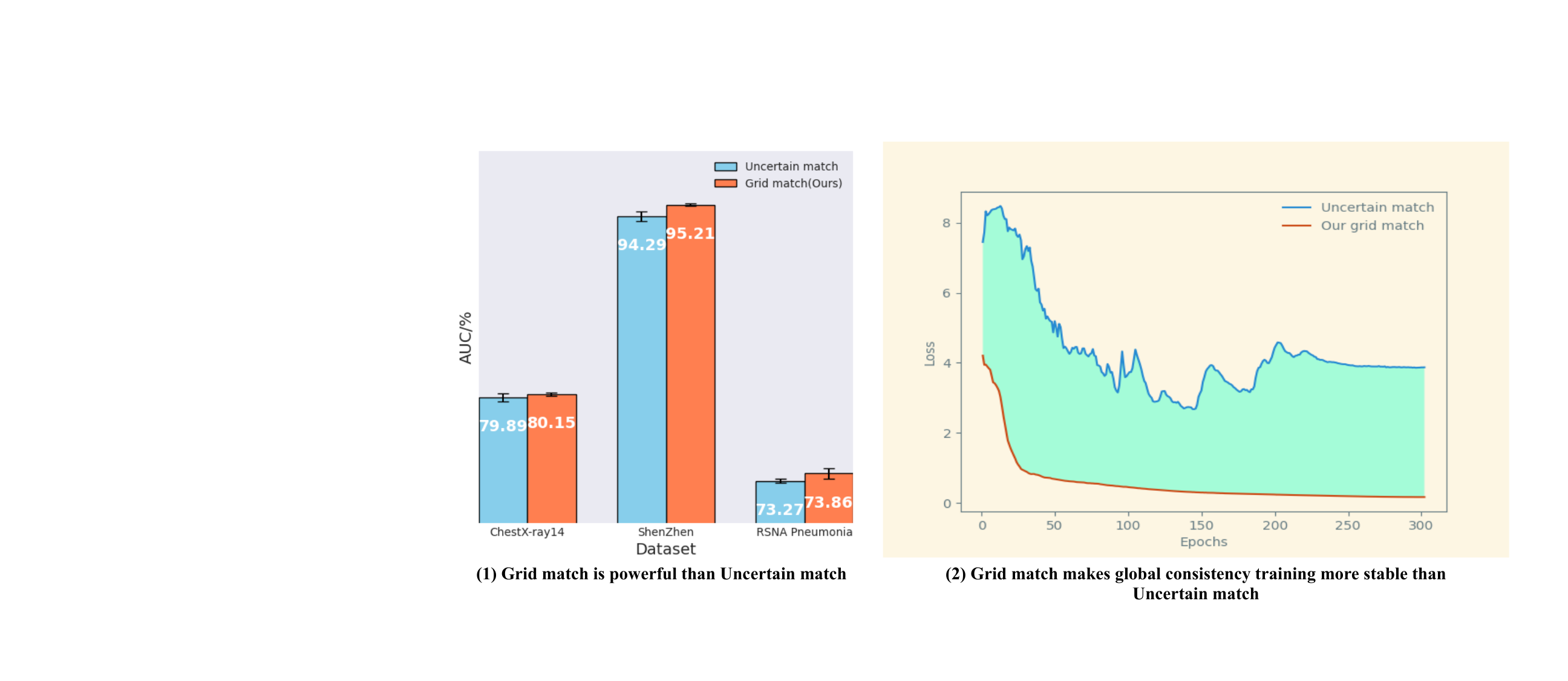}
    \caption{\jlnoterem{use spaces properly}{(1) Grid-based} matching show better performance than block matching in VICRegL's~\cite{bardes2022vicregl} uncertain match. (2) The cures for the global-consistency loss show more stability when training with grid-based match.}
    \label{gridvsuncertain} 
    \vspace{-0.5cm}
\end{wrapfigure}

\jlsecreplace{
    \textbf{(1) \jlsecreplace{Grid match}{Transformer grid-based matching} is more stable and powerful than \jlsecnote{define this or use a more precise term?}{uncertain match}.} In order to calculate local consistency, the correspondence between the cross-view patches needs to be calculated. Currently block matching is usually uncertain match~\cite{bardes2022vicregl, yun2022patch} derived as adjacent position and cosine similarity of embedding vectors between blocks, leading to incomplete and unreliable matches respectively, which cause damage to training stability and reliability while our method is more stable as shown in \figname\ref{gridvsuncertain}.
}{
    \textbf{(1) Transformer grid-based matching is more stable.} Calculating local consistency requires the correspondence between patches across views. Existing block matching~\cite{bardes2022vicregl,yun2022patch} is derived based on the adjacent position or cosine similarity of embedding vectors between blocks, leading to inexact matches, resulting in training instability and unreliability. By contrast, our grid-based matching (illustrated in \figname\ref{fig:cropping}) is more stable as shown in \figname\ref{gridvsuncertain}.

}

\smallskip
\noindent\textbf{(2) Local consistency improves performance.} We remove the local loss and use several different backbones 
to demonstrate the effectiveness of local embedding consistency. The discussions and table of the results can be found in the Appendix.\\

\smallskip
\noindent
\textbf{(3) \jlsecreplace{Teacher-Student}{Student-teacher} model (global consistency) boosts all one branch methods.} In our experiments we found that the \jlsecreplace{teacher-student}{student-teacher} model can be applied to \jlsecdelete{all} one-branch self-supervised methods and boosts their performance. \jlsecreplace{Related results can be found}{The results are included} in the Appendix.


\subsection{Visualization of Upstream Models}\label{sec:upstream}

\begin{figure}[h]
    \centering
    \includegraphics[width=12.5cm]{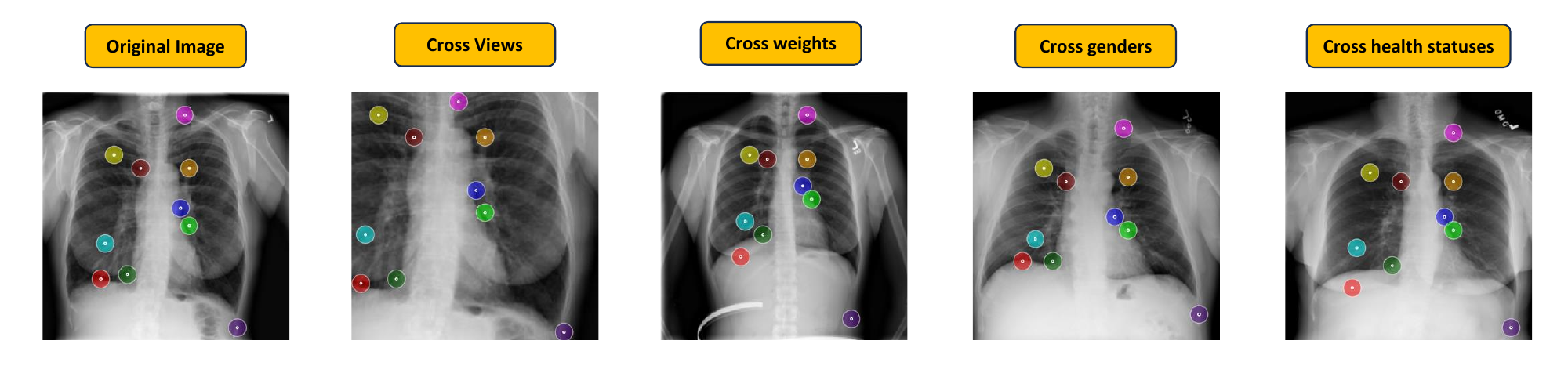}
    \caption{\jlthirdnote{should this figure be replaced with the figure in the appendix? I have try it, but when I do the replacement, the page limit reaches.}{}\jlnoterem{You may change ``Cross Sample'' to ``Cross-patient correspondence'' and ``Cross View'' to ``Cross-view correspondence'' }{}\jlcleandelete{Matching anatomical structures across patients and across views of the same patients.}\jlthirdnoterem{consider matching anatomical structures from a normal patient to a patient with diseases and vice versa -- you may use those X-rays in \figname\ref{fig:motivation}}{} Our PEAC model \jlreplace{effectively identifies}{can effectively \jlthirdreplace{identify}{localize}} \jlthirddelete{the} arbitrary \jlthirddelete{same} anatomical structures across \jlreplace{diverse samples, including those with significant variations such as male and female or fat and thin individuals, as well as between different partially cropped images.}{views of the same patient and across patients of different genders and weights \jlthirdinsert{and of health and disease}.} \jlthirdnoterem{Consider including a figure in the appendix that has 10 patients, each with a few random anatomical structures. For each patient, localize those anatomical structures in different views of the same patient, in distinct patients of the opposite gender, of different weights, and of different healthy statuses.}{} }
    \label{fig4} 
    \label{fig:matching}
\end{figure}

{\bf (1) \jlsecnoterem{make all headlines consistent in terms of capitalization}{Cross-patient and cross-view correspondence}.} 
To show that our PEAC can learn a variety of anatomical structure effectively, we match patch-level features across patients and different views of the same image, as shown in \figname\ref{fig:matching}. Our observations suggest that these features capture semantic regions and exhibit robustness across samples with large morphological differences and both sexes. Similar semantic regions are also stably captured in different views of the same sample. We also compared our results to models trained with other methods, which exhibited obvious incorrect matches. Further details are available in the Appendix.

\begin{figure}[h!]
    \centering
    \includegraphics[width=12.5cm]{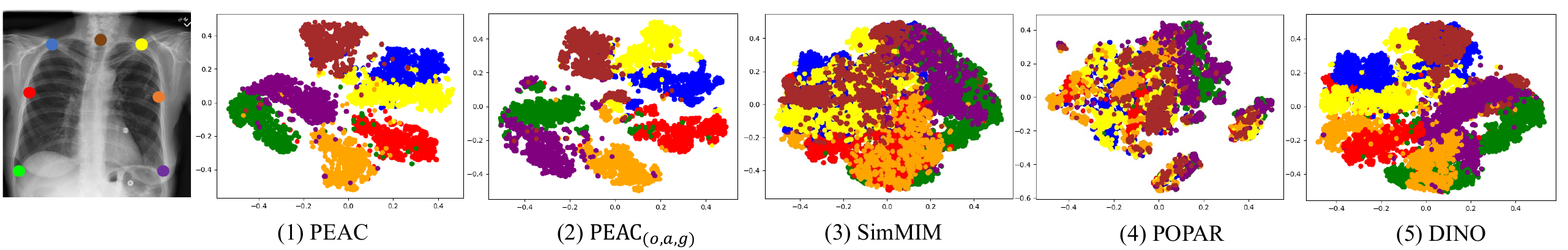}
    \caption{We use t-SNE to contrast our PEAC's landmark embeddings with SimMIM, POPAR, and DINO. PEAC clearly delineates each landmark embedding, demonstrating valid embedding space. However, the reduced \jlsecreplace{PEAC$_(o,a,g)$}{PEAC$_{(o,a,g)}$}, lacking local consistency, shows intermingled landmarks, emphasizing local consistency's role in defining distinct anatomical structure embeddings. \jlthirdnoterem{The locality property of PEAC seems weaker that that of Adam/Eve. Please consider a loss for the dissimilarity between the nonoverlapped regions, that is, we pull the overlapped regions together and push the nonoverlapped regions apart. Also, please start looking into compositionality}{}}
    \label{tsne}
    \vspace{-0.2cm}
\end{figure}

\smallskip
\noindent
{\bf(2) t-SNE of landmark anatomies across patients.} We use t-SNE on 7 labeled local landmark anatomies across 1000 patient images. Each local anatomy is labeled with different color which corresponds with the t-sne cluster color in \figname\ref{tsne}. This shows our PEAC's ability to learn a valid embedding space among different local anatomical structures.

\begin{figure}[h!]
    \centering
    \includegraphics[width=12cm]{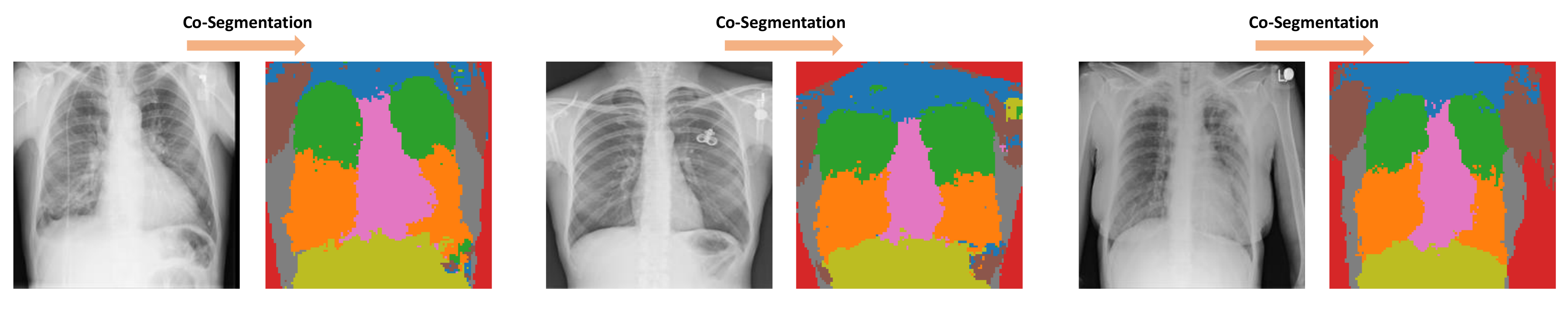}
    \caption{We semantically \jlreplace{co-segments}{co-segment}
common structure of images in a zero-shot scenario. The cervico scapular region, upper lobe of lungs, lower lobe of lungs, mediastinum, and abdominal cavity are clearly segmented as common features.}
    \label{co-segment}
    \vspace{-0.3cm}
\end{figure}

\smallskip
\noindent 
{\bf(3) \jlsecnoterem{make all headlines consistent in terms of capitalization}{Zero-shot co-segmentation}.}  Without finetuning on downstream tasks, we jointly segment analogous structures common to all images in a given set as shown in \figname\ref{co-segment}. 
In our segmentation results, different anatomical structures clearly segmented as common features suggests that our proposed model demonstrates proficiency in extracting and representing the distinguishing features of various anatomical regions.
\section{Conclusion}
We propose a novel self-supervised learning approach, denoted as PEAC, designed to enhance the consistency in learning visual representations of anatomical structures within medical images. The vital technique in PEAC is our novel yet reliable grid-based matching which guarantees both global and local consistency in anatomy. Through \jldelete{the} extensive experiments, we demonstrate the effectiveness of our \jlnoterem{use spaces properly}{scheme. By} accurately identifying the features of each common region across patients of different genders and weights and across different views of the same patients, PEAC exhibits a heightened potential for enhanced AI in medical image \jlreplace{analysis}{analysis.} 

\section{Acknowledgement}
In an earlier BMVC version, we forgot to cite the method proposed by \citet{amir2021deep} for co-segmentation, which we used in Section~\ref{sec:upstream} to visualize various anatomical regions emergent from the PEAC feature space. This research has been supported in part by ASU and Mayo Clinic through a Seed Grant and an Innovation Grant, and in part by the NIH under Award Number R01HL128785. The content is solely the responsibility of the authors and does not necessarily represent the official views of the NIH. This work has utilized the GPUs provided in part by the ASU Research Computing and in part by the Bridges-2 at Pittsburgh Supercomputing Center through allocation BCS190015 and the Anvil at Purdue University through allocation MED220025 from the Advanced Cyberinfrastructure Coordination Ecosystem: Services \& Support (ACCESS) program, which is supported by National Science Foundation grants \#2138259, \#2138286, \#2138307, \#2137603, and \#2138296. The content of this paper is covered by patents pending.

\bibliography{egbib}

\begin{thebibliography}{43}
\providecommand{\natexlab}[1]{#1}
\providecommand{\url}[1]{\texttt{#1}}
\expandafter\ifx\csname urlstyle\endcsname\relax
  \providecommand{\doi}[1]{doi: #1}\else
  \providecommand{\doi}{doi: \begingroup \urlstyle{rm}\Url}\fi

\bibitem[rsn()]{rsna}
\url{https://www.kaggle.com/c/rsna-pneumonia-detection-challenge}.
\newblock RSNA pneumonia detection challenge (2018).

\bibitem[Amir et~al.(2021)Amir, Gandelsman, Bagon, and Dekel]{amir2021deep}
Shir Amir, Yossi Gandelsman, Shai Bagon, and Tali Dekel.
\newblock Deep vit features as dense visual descriptors.
\newblock \emph{arXiv preprint arXiv:2112.05814}, 2\penalty0 (3):\penalty0 4, 2021.

\bibitem[Bao et~al.(2021)Bao, Dong, Piao, and Wei]{bao2021beit}
Hangbo Bao, Li~Dong, Songhao Piao, and Furu Wei.
\newblock Beit: Bert pre-training of image transformers.
\newblock \emph{arXiv preprint arXiv:2106.08254}, 2021.

\bibitem[Bardes et~al.(2022{\natexlab{a}})Bardes, Ponce, and LeCun]{bardes2022variance}
Adrien Bardes, Jean Ponce, and Yann LeCun.
\newblock Variance-invariance-covariance regularization for self-supervised learning.
\newblock \emph{ICLR, Vicreg}, 2022{\natexlab{a}}.

\bibitem[Bardes et~al.(2022{\natexlab{b}})Bardes, Ponce, and LeCun]{bardes2022vicregl}
Adrien Bardes, Jean Ponce, and Yann LeCun.
\newblock Vicregl: Self-supervised learning of local visual features.
\newblock \emph{arXiv preprint arXiv:2210.01571}, 2022{\natexlab{b}}.

\bibitem[Caron et~al.(2021)Caron, Touvron, Misra, J{\'e}gou, Mairal, Bojanowski, and Joulin]{caron2021emerging}
Mathilde Caron, Hugo Touvron, Ishan Misra, Herv{\'e} J{\'e}gou, Julien Mairal, Piotr Bojanowski, and Armand Joulin.
\newblock Emerging properties in self-supervised vision transformers.
\newblock In \emph{Proceedings of the IEEE/CVF international conference on computer vision}, pages 9650--9660, 2021.

\bibitem[Chen and He(2021)]{chen2021exploring}
Xinlei Chen and Kaiming He.
\newblock Exploring simple siamese representation learning.
\newblock In \emph{Proceedings of the IEEE/CVF conference on computer vision and pattern recognition}, pages 15750--15758, 2021.

\bibitem[Chen et~al.(2020)Chen, Fan, Girshick, and He]{chen2020improved}
Xinlei Chen, Haoqi Fan, Ross Girshick, and Kaiming He.
\newblock Improved baselines with momentum contrastive learning.
\newblock \emph{arXiv preprint arXiv:2003.04297}, 2020.

\bibitem[Chen et~al.(2021)Chen, Xie, and He]{chen2021empirical}
Xinlei Chen, Saining Xie, and Kaiming He.
\newblock An empirical study of training self-supervised vision transformers.
\newblock In \emph{Proceedings of the IEEE/CVF International Conference on Computer Vision}, pages 9640--9649, 2021.

\bibitem[Deng et~al.(2009)Deng, Dong, Socher, Li, Li, and Fei-Fei]{deng2009imagenet}
Jia Deng, Wei Dong, Richard Socher, Li-Jia Li, Kai Li, and Li~Fei-Fei.
\newblock Imagenet: A large-scale hierarchical image database.
\newblock In \emph{2009 IEEE conference on computer vision and pattern recognition}, pages 248--255. Ieee, 2009.

\bibitem[Dosovitskiy et~al.(2020)Dosovitskiy, Beyer, Kolesnikov, Weissenborn, Zhai, Unterthiner, Dehghani, Minderer, Heigold, Gelly, et~al.]{dosovitskiy2020image}
Alexey Dosovitskiy, Lucas Beyer, Alexander Kolesnikov, Dirk Weissenborn, Xiaohua Zhai, Thomas Unterthiner, Mostafa Dehghani, Matthias Minderer, Georg Heigold, Sylvain Gelly, et~al.
\newblock An image is worth 16x16 words: Transformers for image recognition at scale.
\newblock \emph{arXiv preprint arXiv:2010.11929}, 2020.

\bibitem[Haghighi et~al.(2022)Haghighi, Taher, Gotway, and Liang]{haghighi2022dira}
Fatemeh Haghighi, Mohammad Reza~Hosseinzadeh Taher, Michael~B Gotway, and Jianming Liang.
\newblock Dira: Discriminative, restorative, and adversarial learning for self-supervised medical image analysis.
\newblock In \emph{Proceedings of the IEEE/CVF Conference on Computer Vision and Pattern Recognition}, pages 20824--20834, 2022.

\bibitem[He et~al.(2020)He, Fan, Wu, Xie, and Girshick]{he2020momentum}
Kaiming He, Haoqi Fan, Yuxin Wu, Saining Xie, and Ross Girshick.
\newblock Momentum contrast for unsupervised visual representation learning.
\newblock In \emph{Proceedings of the IEEE/CVF conference on computer vision and pattern recognition}, pages 9729--9738, 2020.

\bibitem[He et~al.(2022)He, Chen, Xie, Li, Doll{\'a}r, and Girshick]{he2022masked}
Kaiming He, Xinlei Chen, Saining Xie, Yanghao Li, Piotr Doll{\'a}r, and Ross Girshick.
\newblock Masked autoencoders are scalable vision learners.
\newblock In \emph{Proceedings of the IEEE/CVF Conference on Computer Vision and Pattern Recognition}, pages 16000--16009, 2022.

\bibitem[Henaff(2020)]{henaff2020data}
Olivier Henaff.
\newblock Data-efficient image recognition with contrastive predictive coding.
\newblock In \emph{International conference on machine learning}, pages 4182--4192. PMLR, 2020.

\bibitem[Hjelm et~al.(2018)Hjelm, Fedorov, Lavoie-Marchildon, Grewal, Bachman, Trischler, and Bengio]{hjelm2018learning}
R~Devon Hjelm, Alex Fedorov, Samuel Lavoie-Marchildon, Karan Grewal, Phil Bachman, Adam Trischler, and Yoshua Bengio.
\newblock Learning deep representations by mutual information estimation and maximization.
\newblock \emph{arXiv preprint arXiv:1808.06670}, 2018.

\bibitem[Irvin et~al.(2019)Irvin, Rajpurkar, Ko, Yu, Ciurea-Ilcus, Chute, Marklund, Haghgoo, Ball, Shpanskaya, et~al.]{irvin2019chexpert}
Jeremy Irvin, Pranav Rajpurkar, Michael Ko, Yifan Yu, Silviana Ciurea-Ilcus, Chris Chute, Henrik Marklund, Behzad Haghgoo, Robyn Ball, Katie Shpanskaya, et~al.
\newblock Chexpert: A large chest radiograph dataset with uncertainty labels and expert comparison.
\newblock In \emph{Proceedings of the AAAI conference on artificial intelligence}, volume~33, pages 590--597, 2019.

\bibitem[Jaeger et~al.(2014)Jaeger, Candemir, Antani, W{\'a}ng, Lu, and Thoma]{jaeger2014two}
Stefan Jaeger, Sema Candemir, Sameer Antani, Y{\`\i}-Xi{\'a}ng~J W{\'a}ng, Pu-Xuan Lu, and George Thoma.
\newblock Two public chest x-ray datasets for computer-aided screening of pulmonary diseases.
\newblock \emph{Quantitative imaging in medicine and surgery}, 4\penalty0 (6):\penalty0 475, 2014.

\bibitem[Jing and Tian(2020)]{jing2020self}
Longlong Jing and Yingli Tian.
\newblock Self-supervised visual feature learning with deep neural networks: A survey.
\newblock \emph{IEEE transactions on pattern analysis and machine intelligence}, 43\penalty0 (11):\penalty0 4037--4058, 2020.

\bibitem[Lee et~al.(2021)Lee, Arnab, Guadarrama, Canny, and Fischer]{lee2021compressive}
Kuang-Huei Lee, Anurag Arnab, Sergio Guadarrama, John Canny, and Ian Fischer.
\newblock Compressive visual representations.
\newblock \emph{Advances in Neural Information Processing Systems}, 34:\penalty0 19538--19552, 2021.

\bibitem[Li et~al.(2020)Li, Zhou, Xiong, and Hoi]{li2020prototypical}
Junnan Li, Pan Zhou, Caiming Xiong, and Steven~CH Hoi.
\newblock Prototypical contrastive learning of unsupervised representations.
\newblock \emph{arXiv preprint arXiv:2005.04966}, 2020.

\bibitem[Liu et~al.(2021)Liu, Lin, Cao, Hu, Wei, Zhang, Lin, and Guo]{liu2021swin}
Ze~Liu, Yutong Lin, Yue Cao, Han Hu, Yixuan Wei, Zheng Zhang, Stephen Lin, and Baining Guo.
\newblock Swin transformer: Hierarchical vision transformer using shifted windows.
\newblock In \emph{Proceedings of the IEEE/CVF international conference on computer vision}, pages 10012--10022, 2021.

\bibitem[Loshchilov and Hutter(2017)]{loshchilov2017decoupled}
Ilya Loshchilov and Frank Hutter.
\newblock Decoupled weight decay regularization.
\newblock \emph{arXiv preprint arXiv:1711.05101}, 2017.

\bibitem[Ma et~al.(2022)Ma, Hosseinzadeh~Taher, Pang, Islam, Haghighi, Gotway, and Liang]{ma2022benchmarking}
DongAo Ma, Mohammad~Reza Hosseinzadeh~Taher, Jiaxuan Pang, Nahid~UI Islam, Fatemeh Haghighi, Michael~B Gotway, and Jianming Liang.
\newblock Benchmarking and boosting transformers for medical image classification.
\newblock In \emph{Domain Adaptation and Representation Transfer: 4th MICCAI Workshop, DART 2022, Held in Conjunction with MICCAI 2022, Singapore, September 22, 2022, Proceedings}, pages 12--22. Springer, 2022.

\bibitem[Oord et~al.(2018)Oord, Li, and Vinyals]{oord2018representation}
Aaron van~den Oord, Yazhe Li, and Oriol Vinyals.
\newblock Representation learning with contrastive predictive coding.
\newblock \emph{arXiv preprint arXiv:1807.03748}, 2018.

\bibitem[Oron et~al.(2017)Oron, Dekel, Xue, Freeman, and Avidan]{oron2017best}
Shaul Oron, Tali Dekel, Tianfan Xue, William~T Freeman, and Shai Avidan.
\newblock Best-buddies similarity—robust template matching using mutual nearest neighbors.
\newblock \emph{IEEE transactions on pattern analysis and machine intelligence}, 40\penalty0 (8):\penalty0 1799--1813, 2017.

\bibitem[Pang et~al.(2022)Pang, Haghighi, Ma, Islam, Hosseinzadeh~Taher, Gotway, and Liang]{pang2022popar}
Jiaxuan Pang, Fatemeh Haghighi, DongAo Ma, Nahid~Ul Islam, Mohammad~Reza Hosseinzadeh~Taher, Michael~B Gotway, and Jianming Liang.
\newblock Popar: Patch order prediction and appearance recovery for self-supervised medical image analysis.
\newblock In \emph{Domain Adaptation and Representation Transfer: 4th MICCAI Workshop, DART 2022, Held in Conjunction with MICCAI 2022, Singapore, September 22, 2022, Proceedings}, pages 77--87. Springer, 2022.

\bibitem[Shiraishi et~al.(2000)Shiraishi, Katsuragawa, Ikezoe, Matsumoto, Kobayashi, Komatsu, Matsui, Fujita, Kodera, and Doi]{shiraishi2000development}
Junji Shiraishi, Shigehiko Katsuragawa, Junpei Ikezoe, Tsuneo Matsumoto, Takeshi Kobayashi, Ken-ichi Komatsu, Mitate Matsui, Hiroshi Fujita, Yoshie Kodera, and Kunio Doi.
\newblock Development of a digital image database for chest radiographs with and without a lung nodule: receiver operating characteristic analysis of radiologists' detection of pulmonary nodules.
\newblock \emph{American Journal of Roentgenology}, 174\penalty0 (1):\penalty0 71--74, 2000.

\bibitem[Tajbakhsh et~al.(2021)Tajbakhsh, Roth, Terzopoulos, and Liang]{Tajbakhsh2021a}
Nima Tajbakhsh, Holger Roth, Demetri Terzopoulos, and Jianming Liang.
\newblock {Guest Editorial Annotation-Efficient Deep Learning: The Holy Grail of Medical Imaging}.
\newblock \emph{IEEE Transactions on Medical Imaging}, 40\penalty0 (10):\penalty0 2526--2533, oct 2021.
\newblock ISSN 1558254X.
\newblock \doi{10.1109/TMI.2021.3089292}.

\bibitem[Taleb et~al.(2021)Taleb, Lippert, Klein, and Nabi]{taleb2021multimodal}
Aiham Taleb, Christoph Lippert, Tassilo Klein, and Moin Nabi.
\newblock Multimodal self-supervised learning for medical image analysis.
\newblock In \emph{Information Processing in Medical Imaging: 27th International Conference, IPMI 2021, Virtual Event, June 28--June 30, 2021, Proceedings}, pages 661--673. Springer, 2021.

\bibitem[Tarvainen and Valpola(2017)]{tarvainen2017mean}
Antti Tarvainen and Harri Valpola.
\newblock Mean teachers are better role models: Weight-averaged consistency targets improve semi-supervised deep learning results.
\newblock \emph{Advances in neural information processing systems}, 30, 2017.

\bibitem[Tian et~al.(2020)Tian, Krishnan, and Isola]{tian2020contrastive}
Yonglong Tian, Dilip Krishnan, and Phillip Isola.
\newblock Contrastive multiview coding.
\newblock In \emph{Computer Vision--ECCV 2020: 16th European Conference, Glasgow, UK, August 23--28, 2020, Proceedings, Part XI 16}, pages 776--794. Springer, 2020.

\bibitem[Wang et~al.(2017)Wang, Peng, Lu, Lu, Bagheri, and Summers]{wang2017chestx}
Xiaosong Wang, Yifan Peng, Le~Lu, Zhiyong Lu, Mohammadhadi Bagheri, and Ronald~M Summers.
\newblock Chestx-ray8: Hospital-scale chest x-ray database and benchmarks on weakly-supervised classification and localization of common thorax diseases.
\newblock In \emph{Proceedings of the IEEE conference on computer vision and pattern recognition}, pages 2097--2106, 2017.

\bibitem[Wang et~al.(2021)Wang, Zhang, Shen, Kong, and Li]{wang2021dense}
Xinlong Wang, Rufeng Zhang, Chunhua Shen, Tao Kong, and Lei Li.
\newblock Dense contrastive learning for self-supervised visual pre-training.
\newblock In \emph{Proceedings of the IEEE/CVF Conference on Computer Vision and Pattern Recognition}, pages 3024--3033, 2021.

\bibitem[Xiao et~al.(2018)Xiao, Liu, Zhou, Jiang, and Sun]{xiao2018unified}
Tete Xiao, Yingcheng Liu, Bolei Zhou, Yuning Jiang, and Jian Sun.
\newblock Unified perceptual parsing for scene understanding.
\newblock In \emph{Proceedings of the European conference on computer vision (ECCV)}, pages 418--434, 2018.

\bibitem[Xiao et~al.(2021)Xiao, Reed, Wang, Keutzer, and Darrell]{xiao2021region}
Tete Xiao, Colorado~J Reed, Xiaolong Wang, Kurt Keutzer, and Trevor Darrell.
\newblock Region similarity representation learning.
\newblock In \emph{Proceedings of the IEEE/CVF International Conference on Computer Vision}, pages 10539--10548, 2021.

\bibitem[Xie et~al.(2021)Xie, Lin, Zhang, Cao, Lin, and Hu]{xie2021propagate}
Zhenda Xie, Yutong Lin, Zheng Zhang, Yue Cao, Stephen Lin, and Han Hu.
\newblock Propagate yourself: Exploring pixel-level consistency for unsupervised visual representation learning.
\newblock In \emph{Proceedings of the IEEE/CVF Conference on Computer Vision and Pattern Recognition}, pages 16684--16693, 2021.

\bibitem[Xie et~al.(2022)Xie, Zhang, Cao, Lin, Bao, Yao, Dai, and Hu]{xie2022simmim}
Zhenda Xie, Zheng Zhang, Yue Cao, Yutong Lin, Jianmin Bao, Zhuliang Yao, Qi~Dai, and Han Hu.
\newblock Simmim: A simple framework for masked image modeling.
\newblock In \emph{Proceedings of the IEEE/CVF Conference on Computer Vision and Pattern Recognition}, pages 9653--9663, 2022.

\bibitem[Yi and Yoon(2020)]{yi2020patch}
Jihun Yi and Sungroh Yoon.
\newblock Patch svdd: Patch-level svdd for anomaly detection and segmentation.
\newblock In \emph{Proceedings of the Asian Conference on Computer Vision}, 2020.

\bibitem[Yun et~al.(2022)Yun, Lee, Kim, and Shin]{yun2022patch}
Sukmin Yun, Hankook Lee, Jaehyung Kim, and Jinwoo Shin.
\newblock Patch-level representation learning for self-supervised vision transformers.
\newblock In \emph{Proceedings of the IEEE/CVF Conference on Computer Vision and Pattern Recognition}, pages 8354--8363, 2022.

\bibitem[Zbontar et~al.(2021)Zbontar, Jing, Misra, LeCun, and Deny]{zbontar2021barlow}
Jure Zbontar, Li~Jing, Ishan Misra, Yann LeCun, and St{\'e}phane Deny.
\newblock Barlow twins: Self-supervised learning via redundancy reduction.
\newblock In \emph{International Conference on Machine Learning}, pages 12310--12320. PMLR, 2021.

\bibitem[Zhou et~al.(2017)Zhou, Shin, Zhang, Gurudu, Gotway, and Liang]{zhou2017fine}
Zongwei Zhou, Jae Shin, Lei Zhang, Suryakanth Gurudu, Michael Gotway, and Jianming Liang.
\newblock Fine-tuning convolutional neural networks for biomedical image analysis: actively and incrementally.
\newblock In \emph{Proceedings of the IEEE conference on computer vision and pattern recognition}, pages 7340--7351, 2017.

\bibitem[Zhou et~al.(2021)Zhou, Sodha, Pang, Gotway, and Liang]{zhou2021models}
Zongwei Zhou, Vatsal Sodha, Jiaxuan Pang, Michael~B Gotway, and Jianming Liang.
\newblock Models genesis.
\newblock \emph{Medical image analysis}, 67:\penalty0 101840, 2021.

\end{thebibliography}

\clearpage

\appendix
\section*{Appendix}

\section{Experiments}
\subsection{Pretraining settings}

\jlthirdreplace{
    \jlthirdnote{pretty good writing}{\jpnote{Due to notable scalability, global receptibility, and interpretability~\cite{dosovitskiy2020image,caron2021emerging} exhibited by vision transformer architectures, we implement PEAC on two mainstream vision transformer architectures, namely ViT-B and Swin-B. To train PEAC, we have utilized the ChestX-ray14 dataset and amalgamate the official training and validation splits to serve as our training set. For both architectures, ViT-B and Swin-B, the input image size is set to 224×224, which leads to 196 (14×14) shufflable patches in the case of ViT-B. The Swin-B, on the other hand, resulting in 49 (7×7) shufflable patches due to its hierarchical structure. To learn the same contextual relationship as the ViT-B, we pretrain PEAC on Swin-B with 448×448 input image size, while the tissue (physical) size remains unchanged, resulting in the same 196 (14 × 14) shufflable patches.}{}}
}{
    We implement PEAC on ViT-B~\cite{dosovitskiy2020image} and Swin-B~\cite{liu2021swin} for their notable scalability, global receptibility, and interpretability~\cite{dosovitskiy2020image,caron2021emerging}. Both PEACs are trained on ChestX-ray14~\cite{wang2017chestx} by amalgamating the official training and validation splits. In PEAC ViT-B, input images of size 224×224 lead to 196 (14×14) shufflable patches, while in PEAC Swin-B, it results in 49 (7×7) shufflable patches due to the Swin hierarchical architecture. To learn the same contextual relationship as in PEAC ViT-B, we pretrain PEAC Swin-B with images of size 448×448, but the tissue (physical) size covered by the images remains unchanged, resulting in the same 196 (14 × 14) shufflable patches in terms of the (physical) tissue size.
}

\jlthirdreplace{
    \smallskip
    \noindent
    \jlthirdnote{write in formal English; wordy, too}{
    In the construction of our model, a multi-class linear layer is designated for patch order classification, while a single convolutional block is employed for patch appearance restoration. The global and local consistency branches make use of two 3-layer MLPs as expanders before the computation of consistency losses. The model's learning parameters include a learning rate of 0.1, a momentum of 0.9 for the SGD optimizer, and a warmup period of 5 epochs. The teacher model's parameters are updated post each iteration, with an EMA updating parameter set at 0.999, and a batch size of 8. We utilized four Nvidia RTX3090 GPUs for training PEAC models with an image size of 224$\times$224 for 300 epochs, but we reduce the number of epoch to 150 when training size images of 448$\times$448.
    }
}{
    In PEAC, a multi-class linear layer is designated for patch order classification (\eqname\ref{eq:poc}), and a single convolutional block is employed for patch appearance restoration (\eqname\ref{eq:par}). The global and local consistency branches utilize two 3-layer MLPs as expanders before computing consistency losses. When training PEAC, we use a learning rate of 0.1, a momentum of 0.9 for the SGD optimizer, a warmup period of 5 epochs, and a batch size of 8. The teacher model is updated after each iteration via EMA with an updating parameter of 0.999. We utilize four Nvidia RTX3090 GPUs for training PEAC models with images of size 224$\times$224 for 300 epochs, but we reduce the number of epoch to 150 when the image size is 448$\times$448.
}

\subsection{Target Tasks and Datasets}

We \jlthirdreplace{fine-tune}{evaluate} our PEAC models \jlthirdreplace{to}{by finetuning on} four \jlthirdnote{no segmentation target tasks? The power of PEAC is for segmentation and possible registration}{classification} target tasks ChestX-ray14~\cite{wang2017chestx}, CheXpert~\cite{irvin2019chexpert}, NIH Shenzhen CXR~\cite{jaeger2014two}, RSNA Pneumonia~\cite{rsna} and one segmentation task JSRT~\cite{shiraishi2000development}: \jlthirdnoterem{When itemizing/listing, use consistent grammatical structures}{} \jlthirdnoterem{if no page limit, please itemize the datasets.}{
\begin{itemize}
    \item \jlthirdnoterem{make this consistent across all items}{{\bf ChestX-ray14}~\cite{wang2017chestx}, which contains 112K frontal-view X-ray images of 30805 unique patients with the text-mined fourteen disease image labels (where each image can have multi-labels).} We use the \jlthirdreplace{offcial}{official} training set 86K (90\% for training and 10\% for validation) and testing set 25K.

    \item \jlthirdnoterem{make this consistent across all items}{{\bf CheXpert}~\cite{irvin2019chexpert}, which includes 224K chest radiographs of 65240 patients and capturing 14 thoracic diseases.} We use \jlthirdnote{This is not the official split -- please check with Jiaxuan for the official test dataset released recently -- use the official test dataset in the future}{the official training data split 224K for training and validation set 234 images for testing}. We train on the 14 thoracic diseases but \jlthirdreplace{we test only}{follow the standard practice by testing} on 5 diseases (Atelectasis, Cardiomegaly, Consolidation, Edema, and Pleural Effusion).
    
    \item \jlthirdnoterem{make this consistent across all items}{{\bf NIH Shenzhen CXR}~\cite{jaeger2014two}, which contains 326 normal and 336 Tuberculosis (TB) frontal-view chest X-ray images.} We split 70\% of the dataset for training, 10\% for validation and 20\% for testing which are the same with ~\cite{ma2022benchmarking};
    
    \item \jlthirdnoterem{make this consistent across all items}{{\bf RSNA Pneumonia}~\cite{rsna}, which consists of 26.7K frontal view chest X-ray images and each image is labeled with a distinct diagnosis, such as Normal, Lung Opacity and Not Normal (other diseases).} 80\% of the images are used to train, 10\% to valid and 10\% to test. These target datasets are composed of both multi-label and multi-class classification tasks with various diseases.

    \item \textbf{JSRT}~\cite{shiraishi2000development}, which is a organ segmentation dataset including 247 frontal view chest X-ray images. All of them are in 2048$\times$2048 resolution with 12-bit gray-scale levels. Both lung, heart and clavicle segmentation masks are available for this dataset. We split 173 images for training, 25 for validation and 49 for testing.

\end{itemize}
}

\subsection{Finetuning Setting}
\setcounter{figure}{7}
\begin{figure}[h]
    \centering
    \includegraphics[width=6.0cm]{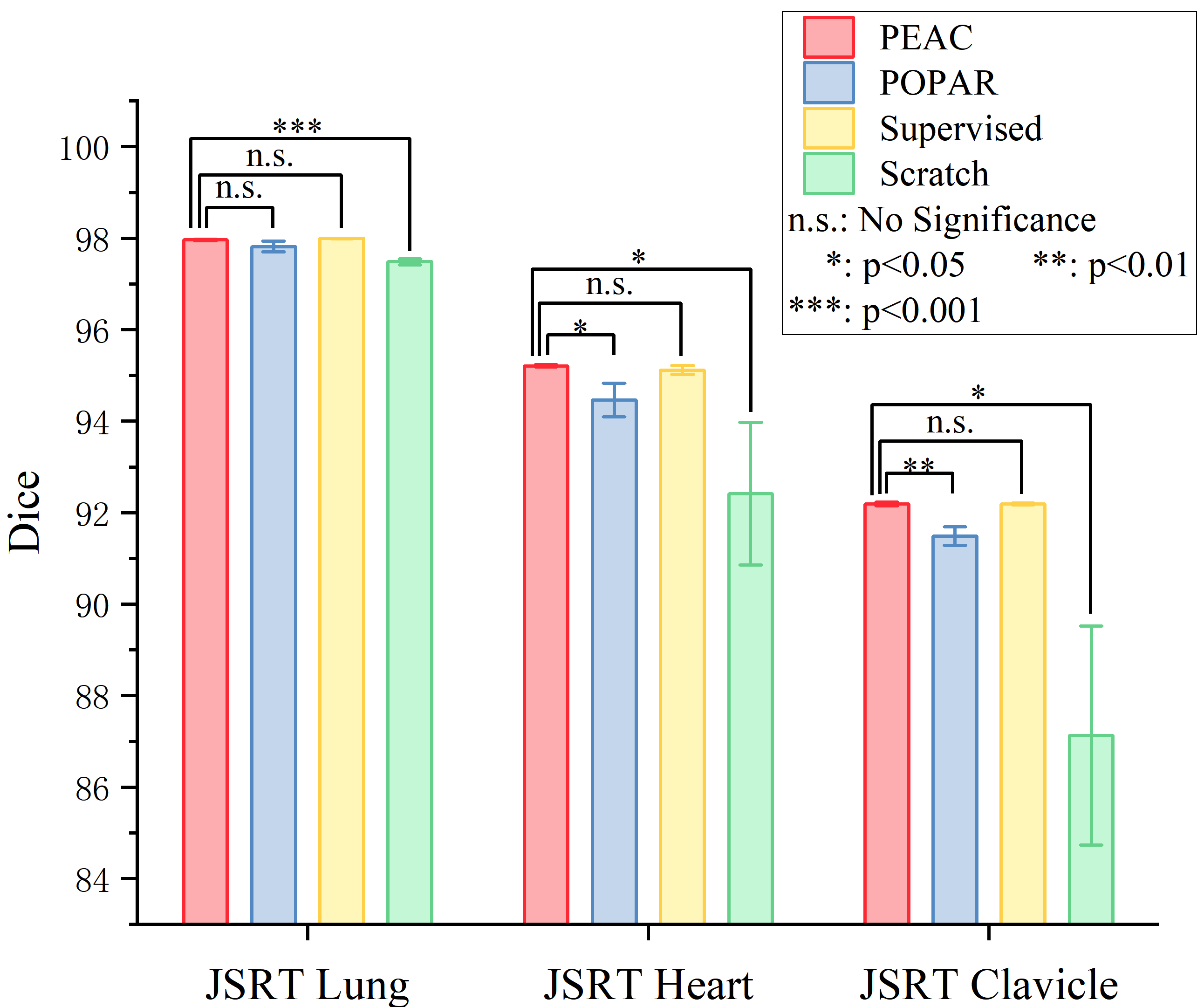}
    \caption{Comparison of segmentation results on JSRT dataset. To investigate the performance on segmentation tasks, we compare PEAC with SSL method POPAR, fully-supervised model, and model training from scratch.}
    \label{seg_pvalue}
\end{figure}
\jlthirdnote{Wordy}{We transfer the PEAC pretrained models to each target task by fine-tuning \jlthirdnote{wordy}{the whole parameters for the target classification tasks}. For the target classification tasks, we concatenate a randomly initialized linear layer to the output of the classification (CLS) token of PEAC ViT-B models. Due to the structural \jlthirdnote{make the ``difference'' explicitly with whom}{difference with \jlthirdnote{Do you use the CLS token in ViT-B?}{ViT-B model}}, PEAC Swin-B models don't equip the CLS token and we add an average pooling to the last-layer feature maps, then feed the feature to the randomly initialized linear layer. The AUC (area under the ROC curve) is used to evaluate the multi-label classification performance (ChestX-ray14, CheXpert and NIH Shenzhen CXR), while the accuracy is used to assess the multi-class classification performance (RSNA Pneumonia). For the target segmentation task, we use UperNet~\cite{xiao2018unified} as the training model. We concatenate pretrained Swin-B and randomly initialized prediction head for segmenting. In JSRT dataset, we independently train 3 models for the three organs lung, heart and clavicle and the Dice is used to evaluate the segmentation performance, as shown in and \figname\ref{seg_pvalue} .  Following~\cite{liu2021swin}, in fine-tuning experiments we use AdamW~\cite{loshchilov2017decoupled} optimizer with a cosine learning rate scheduler, linear warm up of 20 epochs while the overall epoch is 150, and 0.0005 for the maximum learning rate value. The batch sizes are 32 and 128 for image sizes of 448 and 224, respectively. We train with \jlthirdnote{make the best utilization of 200+ A100 at ASU during the summer}{single Nvidia RTX3090 24G GPU} for performing each experiment.}

\jlthirdnote{The comparison between ViT-B and Swin-B might not be fair then. \textcolor{green}{In the downstream tasks, the CLS token in ViT-B is used for classification.}}{}


\jlthirdreplace{
    \subsection{\jlthirdnoterem{Please check my definitions for some versions -- please let know what you think.}{Definition of the setup name} and the performances }
    (Your definitions are good, but what we have implemented are the versions I filled the results in \tablename\ref{tab:combinations}. What do you think the definition I filled in \tablename\ref{tab:POPAR and PEAC versions} ?)
}{
    \section{Ablation Studies: PEAC versions and their performance}
}

\jlthirdreplace{}{
\noindent
Our PEAC involves four losses:
    \begin{itemize}
    \item Patch order classification loss defined in the main paper as
    \begin{equation}
        \mathcal{L}^{oc}_{\theta_s}=-\frac{1}{B} \sum_{b=1}^B \sum_{l=1}^n \sum_{c=1}^n \mathcal{Y} \log \mathcal{P}^{o}
        \label{eq:poc}
    \end{equation}
    
    \item Patch appearance restoration loss defined in the main paper as
    \begin{equation}
        \mathcal{L}^{ar}_{\theta_s}=\frac{1}{B} \sum_{i=1}^B \sum_{j=1}^n\left\|p_j-p_j^{a}\right\|_2^2
        \label{eq:par}
    \end{equation}

    \item Global patch embedding consistency loss defined in the main paper as $\mathcal{L}^{G}_{\theta_s, \theta_t} = \mathcal{L}^{global}_{\theta_s, \theta_t}+\widetilde{\mathcal{L}}^{global}_{\theta_s, \theta_t}$. $\mathcal{L}^{global}_{\theta_s, \theta_t}$ and $\widetilde{\mathcal{L}}^{global}_{\theta_s, \theta_t}$ are computed from an exchanged inputs into the Student and Teacher models.\jlthirdnoterem{need the loss used in \tablename\ref{tab:POPAR and PEAC versions}}{}
    \begin{equation}
        \mathcal{L}^{global}_{\theta_s, \theta_t} \triangleq \|\overline{y_s} - \overline{y_t}\|_2^2 = 2-2\cdot \frac{y_s-y_t}{\| y_s\|_2 \cdot \|y_t \|_2}
    \end{equation}

    \item Local patch embedding consistency loss defined in the main paper as $\mathcal{L}^{L}_{\theta_s, \theta_t} = \mathcal{L}^{local}_{\theta_s, \theta_t}+\widetilde{\mathcal{L}}^{local}_{\theta_s, \theta_t}$. $\mathcal{L}^{local}_{\theta_s, \theta_t}$ and $\widetilde{\mathcal{L}}^{local}_{\theta_s, \theta_t}$ are computed from an exchanged inputs into the Student and Teacher models. \jlthirdnoterem{need the loss used in \tablename\ref{tab:POPAR and PEAC versions}}{}
    \begin{equation}
        \mathcal{L}^{local}_{\theta_s, \theta_t} \triangleq \frac{1}{B} \sum_{b=1}^B \mathbb{I} \cdot (\sum_{i=1}^z \|\overline{p_{m_i}} - \overline{p_{n_i}}\|_2^2)
    \end{equation}
    \end{itemize}
}


\begin{table}[hbp]
\tiny 
    \centering
    \setlength{\belowcaptionskip}{0.2cm}
    \caption{We add the four loss functions one by one to show the effectiveness of our method in terms of performance. 
    All models in the ablation studies are pretrained on ChestX-ray14~\cite{wang2017chestx} with Swin-B backbone at two different image resolutions, and they are also fine-tuned at two different image resolutions as denoted by PT$\rightarrow$FT in the table. Our official implementation PEAC achieves the best performance or the second best on three target tasks with pretraining and finetuning resolutions set at $448\times 448$. \jlthirdnote{compute p-values and highlight rows as in \figname~\ref{seg results}.}{}
    }
    \label{tab:POPAR and PEAC versions}
    \resizebox{1.0\columnwidth}{!}{
    \begin{tabular}{c|c|c|cc|cc|cc|ccc}
    \toprule
        \multirow{2}*{PEAC Version} & Shuffled & \multirow{2}*{PT$\rightarrow$FT} & \multicolumn{2}{c|}{Transformations} & \multicolumn{2}{c|}{POPAR Losses} & \multicolumn{2}{c|}{PEAC Losses} & \multicolumn{3}{c}{Target Tasks}\\
        \cline{4-12}
         & patches & & OD & AD & $\mathcal{L}^{oc}_{\theta_s}$ & $\mathcal{L}^{ar}_{\theta_s}$ & $\mathcal{L}^{G}_{\theta_s, \theta_t}$ & $\mathcal{L}^{L}_{\theta_s, \theta_t}$ & ChestX-ray14 & ShenZhen & RSNA Pneumonia \\
    \midrule
        PEAC$_{(o)}^{-2}$&\multirow{3}*{49} & \multirow{3}*{$224^{2}\rightarrow 224^{2}$} & \cmark & \xmark &\cmark & \xmark & \xmark & \xmark & 78.58$\pm$0.17 & 92.65$\pm$0.65 & 71.46$\pm$0.41 \\
         PEAC$_{(a)}^{-2}$&&& \xmark & \cmark & \xmark & \cmark & \xmark & \xmark & 79.35$\pm$0.18& 93.85$\pm$0.09 & 72.38$\pm$0.15 \\
         PEAC$_{(o,a)}^{-2}$\jlthirdnote{$\neq$PEAC$_{(od,ad)}^{-2}$$\equiv$POPAR$^{-2})$?}{}&&& \cmark & \cmark & \cmark & \cmark & \xmark & \xmark & 79.57$\pm$0.22 & 95.10$\pm$0.20 & 72.59$\pm$0.13\\
        \midrule
        PEAC$_{(g)}^{-2}$ & \multirow{5}*{49} & \multirow{5}*{$224^{2}\rightarrow 224^{2}$} & \xmark & \xmark &\xmark & \xmark & \cmark & \xmark & $80.85 \pm 0.14$ & $96.59 \pm 0.11$ & $73.42 \pm 0.41$ \\
        PEAC$_{(o,g)}^{-2}$ & & & \cmark & \xmark &\cmark & \xmark & \cmark & \xmark & $81.13 \pm 0.18$ & $96.70 \pm 0.11$ & $73.75 \pm 0.04$ \\
         PEAC$_{(o,g,l)}^{-2}$ && & \cmark & \xmark & \cmark & \xmark & \cmark & \cmark & $81.09 \pm 0.35$ & $97.00 \pm 0.28$ & \cellcolor{cyan!25}$\bm{74.42 \pm 0.34}$\\ 
        PEAC$_{(o,a,g)}^{-2}$ & & & \cmark & \cmark & \cmark & \cmark & \cmark & \xmark & $81.25 \pm 0.16$ & $96.91 \pm 0.07$ & $73.35 \pm 0.19$ \\
        PEAC$^{-2}$\jlthirdnote{$\neq$PEAC$_{(od,ad,g,l)}^{-2}$?}{}&& & \cmark & \cmark & \cmark & \cmark & \cmark & \cmark & $81.38 \pm 0.03$ & $97.14 \pm 0.10$ & $74.19 \pm 0.15$\\
    \midrule
        PEAC$_{(o,a,g)}^{-1}$ &\multirow{2}*{196} & \multirow{2}*{$448^{2}\rightarrow 224^{2}$} & \cmark & \cmark & \cmark & \cmark & \cmark & \xmark & 81.51 $\pm$ 0.22 & $97.07 \pm 0.37$ & 73.63 $\pm$ 0.42 \\
        PEAC$^{-1}$& & & \cmark & \cmark & \cmark & \cmark & \cmark & \cmark & $81.90 \pm 0.15$ & $97.17 \pm 0.42$ & $73.70 \pm 0.48$\\
    \midrule
        PEAC$_{(o,a,g)}$ &\multirow{2}*{196} & \multirow{2}*{$448^{2}\rightarrow 448^{2}$} & \cmark & \cmark & \cmark & \cmark & \cmark & \xmark & $\underline{82.67 \pm 0.11}$ & $\underline{97.15 \pm 0.40}$ & $74.18 \pm 0.52$\\
        PEAC& & & \cmark & \cmark & \cmark & \cmark & \cmark & \cmark & $\bm{82.78 \pm 0.21}$ & $\bm{97.39 \pm 0.19}$ & \cellcolor{cyan!25}$\underline{74.39 \pm 0.66}$ \\
    \bottomrule
    \end{tabular}}
\end{table}

We remove some ingredients from our \jlthirdnote{This is our local lab language; please use a formal scientific term.}{official implementation} PEAC and the results (\tablename\ref{tab:POPAR and PEAC versions}) show the effectiveness of all loss functions. The POPAR versions involve OD (patch order distortion) and AD (patch appearance distortion) which are studied in ~\cite{pang2022popar} and the losses include patch order classification loss $\mathcal{L}^{oc}_{\theta_s}$, patch appearance restoration loss $\mathcal{L}^{ar}_{\theta_s}$. \jlthirdnote{Jiaxuan: Is this how POPAR was implemented?}{The downgraded version PEAC$_{(o)}^{-2}$ only include OD, in this circumstance we only compute $\mathcal{L}^{oc}_{\theta_s}$ and neglect the $\mathcal{L}^{ar}_{\theta_s}$.} Correspondingly, \jlthirdnote{Jiaxuan: Is this how POPAR was implemented?}{only AD is added for the downgraded version PEAC$_{(a)}^{-2}$, in this case we only compute $\mathcal{L}^{ar}_{\theta_s}$ and neglect the $\mathcal{L}^{oc}_{\theta_s}$.} The PEAC versions involve the four loss functions mentioned above. Under the same settings (the same shuffled patches and the same pretraining and fine-tuning resolutions), we added these loss functions one by one, and the downstream tasks performance improve successively shown in \tablename\ref{tab:POPAR and PEAC versions}.

\begin{table}[hbp]
\tiny 
    \centering
    \setlength{\belowcaptionskip}{0.2cm}
    \caption{\jlthirdreplace{We add local consistency loss basing on different methods.}{The local consistency loss in PEAC consistently improves the performance across methods and target tasks. \jlthirdnote{do you have results for SimMIM$_{(l)}$, which is what you really need in this table?: We have not done this experiment. Or we may remove SimMIM?}{}}
    }
    \label{tab:local consistency}
    \resizebox{1.0\columnwidth}{!}{
    \begin{tabular}{cc|cc|cc|cc|ccc}
    \toprule
        \multirow{2}*{Method} & \multirow{2}*{Backbone} & \multicolumn{2}{c|}{Transformations} & \multicolumn{2}{c|}{POPAR Losses} & \multicolumn{2}{c|}{PEAC Losses} & \multicolumn{3}{c}{Target Tasks}\\
        \cline{3-11}
         &  & OD & AD & $\mathcal{L}^{oc}_{\theta_s}$ & $\mathcal{L}^{ar}_{\theta_s}$ & $\mathcal{L}^{G}_{\theta_s, \theta_t}$ & $\mathcal{L}^{L}_{\theta_s, \theta_t}$ & ChestX-ray14 & ShenZhen & RSNA Pneumonia \\
    \midrule
        VICRegL& \multirow{2}*{ConvNeXt-B} & \xmark & \xmark &\xmark & \xmark & \xmark & \xmark & 79.89$\pm$0.34 & 94.29$\pm$0.40 & 73.27$\pm$0.15 \\

         VICRegL$_{(l)}$&& \xmark & \xmark & \xmark & \xmark & \xmark & \cmark & 80.15$\pm$0.11 & 95.21$\pm$0.11 & 73.86$\pm$0.43\\
    \midrule
        SimMIM& \multirow{3}*{Swin-B} & \xmark & \xmark &\xmark & \xmark & \xmark & \xmark & 79.09$\pm$0.57 & 93.03$\pm$0.48 & 71.99$\pm$0.55 \\
         SimMIM$_{(g)}$&& \xmark & \xmark & \xmark & \xmark & \cmark & \xmark & 81.42$\pm$ 0.04 & 97.11$\pm$ 0.26 & 73.95$\pm$ 0.18\\
         SimMIM$_{(g,l)}$&& \xmark & \xmark & \xmark & \xmark & \cmark & \cmark & 81.67$\pm$ 0.04 & 97.86$\pm$ 0.07 & $\underline{74.25\pm 0.24}$\\
    \midrule
        \jlthirdreplace{PEAC$_{(o,a,g)}$}{PEAC$_{(o,a,g)}$} &\multirow{2}*{Swin-B} & \cmark & \cmark & \cmark & \cmark & \cmark & \xmark & $\underline{82.67 \pm 0.11}$ & $\underline{97.15 \pm 0.40}$ & $74.18 \pm 0.52$\\
        PEAC& & \cmark & \cmark & \cmark & \cmark & \cmark & \cmark & $\bm{82.78 \pm 0.21}$ & $\bm{97.39 \pm 0.19}$ & $\bm{74.39 \pm 0.66}$ \\
        
    \bottomrule
    \end{tabular}}
\end{table}

\begin{table}[hbp]
\tiny 
    \centering
    \setlength{\belowcaptionskip}{0.2cm}
    \caption{\jlthirdreplace{We use Teacher-Student model on different methods.}{The global loss in PEAC consistently boosts the performance across methods and target tasks.}
    }
    \label{tab:teacher-student}
    \resizebox{1.0\columnwidth}{!}{
    \begin{tabular}{c|cc|cc|cc|ccc}
    \toprule
        \multirow{2}*{Method}  & \multicolumn{2}{c|}{Transformations} & \multicolumn{2}{c|}{POPAR Losses} & \multicolumn{2}{c|}{PEAC Losses} & \multicolumn{3}{c}{Target Tasks}\\
        \cline{2-10}
         & OD & AD & $\mathcal{L}^{oc}_{\theta_s}$ & $\mathcal{L}^{ar}_{\theta_s}$ & $\mathcal{L}^{G}_{\theta_s, \theta_t}$ & $\mathcal{L}^{L}_{\theta_s, \theta_t}$ & ChestX-ray14 & ShenZhen & RSNA Pneumonia \\
    \midrule
        SimMIM & \xmark & \xmark &\xmark & \xmark & \xmark & \xmark & 79.09$\pm$0.57 & 93.03$\pm$0.48 & 71.99$\pm$0.55 \\
         SimMIM$_{(g)}$ & \xmark & \xmark & \xmark & \xmark & \cmark & \xmark & \cellcolor{cyan!25}$\bm{81.42\pm 0.04}$ & $\bm{97.11\pm 0.26}$ & $\underline{73.95 \pm 0.18}$\\
    \midrule
        POPAR$_{od}^{-2}$ & \cmark & \xmark &\cmark & \xmark & \xmark & \xmark & 78.58$\pm$0.17 & 92.65$\pm$0.65 & 71.46$\pm$0.41 \\
         PEAC$_{(o,g)}^{-2}$ & \cmark & \xmark & \cmark & \xmark & \cmark & \xmark & $81.13 \pm 0.18$ & $96.70 \pm 0.11$ & $73.75 \pm 0.04$\\
    \midrule
        POPAR$^{-2}$ & \cmark & \cmark & \cmark & \cmark & \xmark & \xmark & 79.57$\pm$0.22 & 95.10$\pm$0.20 & 72.59$\pm$0.13\\
        PEAC$_{(o,a,g)}^{-2}$ & \cmark & \cmark & \cmark & \cmark & \cmark & \xmark & \cellcolor{cyan!25}$\underline{81.38 \pm 0.03}$ & $\underline{96.91 \pm 0.10}$ & $\bm{74.19 \pm 0.15}$ \\
    \bottomrule
    \end{tabular}}
\end{table}

Our pretraining and fine-tuning setting include two resolutions 448$\times$448 and 224$\times$224. The downgraded versions PEAC$^{-2}$ contain 49 pretraining shuffled patches and are pretrained and fine-tuned on 224 size of images while the downgraded versions PEAC$^{-1}$ include 196 shuffled patches and are pretrained on 448 and fine-tuned on 224 size of images. And the performances on our official implementation PEAC (pretrained and fine-tuned on 448 images) are the best. To accelerate the training process, we only pretrain two versions PEAC and PEAC$_{(o,a,g)}$ on 448 images.

\jlthirdnote{POPAR has two transformations: (1) patch order distortion (OD) and (2) patch appearance distortion (AD), and two losses: (a) patch order classification $\mathcal{L}^{oc}_{\theta_s}$ (defined as your \eqname\ref{eq1}) and (b) patch appearance recovery $\mathcal{L}^{ar}_{\theta_s}$ (defined as your \eqname\ref{eq2}). 

\begin{itemize}
\item For OD (defined as POPAR$_{od}$), involving both ``misplaced patch order classification'' and ``misplaced patch appearance recovery'', you may compute one or two losses: (1) the loss for misplaced patch order classification $\mathcal{L}^{oc}_{\theta_s}$ (defined as your \eqname\ref{eq1}); (2) the loss for misplaced patch appearance recovery $\mathcal{L}^{ar}_{\theta_s}$ (defined as your \eqname\ref{eq2}). 

\item For AD (defined as POPAR$_{ad}$), involving no patch displacement but only appearance distortion, but you still may compute both losses: (1) the loss for (distorted; never misplaced in this case) patch order classification $\mathcal{L}^{oc}_{\theta_s}$ (defined as your \eqname\ref{eq2}); (2) the loss for distorted patch appearance recovery $\mathcal{L}^{ar}_{\theta_s}$ (defined as your \eqname\ref{eq2}). 
\end{itemize}

This yields 7 sensible combinations from many possible combinations. {\bf Please {\bf precisely} define and include all versions on one table here}

\begin{table}[hbp]
\jlblue
\tiny 
    \centering
    \setlength{\belowcaptionskip}{0.2cm}
    \caption{\jlthirdnote{I discussed this with Haozhe; please develop an accurate definition and notation of clarity for each of the combinations}{7 sensible combinations in POPAR yield 21 sensible PEAC versions -- please define and give a name to each version in this table.} In research, it is more important to decide ``what {\it not} to do'' than ``deciding what to do'', that is, there is no need to do experiments for all versions -- please eliminate those versions that are not so important for experimenting.}
    \label{tab:combinations}
    \begin{tabular}{c|cc|cc|cc|ccc}
    \toprule
        \multirow{2}*{POPAR Version} & \multicolumn{2}{c|}{Transformations} & \multicolumn{2}{c|}{POPAR Losses} & \multicolumn{2}{c|}{PEAC Losses} & \multicolumn{3}{c}{Target Tasks}\\
        \cline{2-10}
         & OD & AD & $\mathcal{L}^{oc}_{\theta_s}$ & $\mathcal{L}^{ar}_{\theta_s}$ & $\mathcal{L}^{G}_{\theta_s, \theta_t}$ & $\mathcal{L}^{L}_{\theta_s, \theta_t}$ & ChestX-ray14 & ShenZhen & RSNA Pneumonia \\
    \midrule
        \jlthirdreplace{POPAR$_{od}^{-2}$}{POPAR$_{o}^{-2}$}& \cmark & \xmark &\cmark & \xmark & \xmark & \xmark & 78.58$\pm$0.17 & 92.65$\pm$0.65 & 71.46$\pm$0.41 \\
         & \cmark & \xmark & \xmark &  \cmark & \xmark & \xmark &  &  & \\
        {\jlthirdred POPAR$_{od}^{-2}$} & \cmark &  \xmark & \cmark & \cmark & \xmark & \xmark &  &  & \\
    \midrule
         & \xmark  & \cmark & \cmark & \xmark & \xmark &  \xmark &  &  &  \\
         \jlthirdreplace{POPAR$_{ad}^{-2}$}{POPAR$_{a}^{-2}$} & \xmark & \cmark & \xmark & \cmark & \xmark & \xmark & 79.35$\pm$0.18& 93.85$\pm$0.09 & 72.38$\pm$0.15 \\
         {\jlthirdred POPAR$_{ad}^{-2}$} & \xmark & \cmark & \cmark & \cmark & \xmark & \xmark &  &  &  \\ 
    \midrule
         POPAR$^{-2}$\jlthirdinsert{($\equiv$POPAR$_{(od,ad)}^{-2}$)}& \cmark & \cmark & \cmark & \cmark & \xmark & \xmark & 79.57$\pm$0.22 & 95.10$\pm$0.20 & 72.59$\pm$0.13\\
    \bottomrule
    \end{tabular}   
    \begin{tabular}{c|cc|cc|cc|ccc}
    \toprule
        \multirow{2}*{PEAC Version} & \multicolumn{2}{c|}{Transformations} & \multicolumn{2}{c|}{POPAR Losses} & \multicolumn{2}{c|}{PEAC Losses} & \multicolumn{3}{c}{Target Tasks}\\
        \cline{2-10}
         & OD & AD & $\mathcal{L}^{oc}_{\theta_s}$ & $\mathcal{L}^{ar}_{\theta_s}$ & $\mathcal{L}^{G}_{\theta_s, \theta_t}$ & $\mathcal{L}^{L}_{\theta_s, \theta_t}$ & ChestX-ray14 & ShenZhen & RSNA Pneumonia \\
    \midrule
        PEAC$_{(o,g)}^{-2}$& \cmark & \xmark &\cmark & \xmark & \cmark & \xmark & $81.13 \pm 0.18$ & $96.70 \pm 0.11$ & $73.75 \pm 0.04$ \\
        & \cmark & \xmark & \xmark &  \cmark& \cmark & \xmark &  &  & \\
        {\jlthirdred PEAC$_{(od,g)}^{-2}$} & \cmark &  \xmark & \cmark & \cmark & \cmark &  \xmark &  &  & \\
        \cdashline{2-10}
        & \cmark & \xmark &\cmark & \xmark & \xmark & \cmark &  & &   \\
        & \cmark & \xmark & \xmark &  \cmark& \xmark & \cmark &  &  & \\
        {\jlthirdred PEAC$_{(od,l)}^{-2}$} & \cmark &  \xmark & \cmark & \cmark & \xmark & \cmark &  &  & \\
        \cdashline{2-10}
        PEAC$_{(o,g,l)}^{-2}$ & \cmark & \xmark &\cmark & \xmark & \cmark & \cmark &81.09±0.35  &97.00±0.28
 &74.42±0.34   \\
        & \cmark & \xmark & \xmark &  \cmark& \cmark & \cmark &  &  & \\
        {\jlthirdred PEAC$_{od}^{-2}$($\equiv$PEAC$_{(od,g,l)}^{-2}$)}& \cmark &  \xmark & \cmark & \cmark & \cmark & \cmark &  &  & \\
    \midrule
         & \xmark  & \cmark & \cmark & \xmark & \cmark & \xmark &  &  &  \\
         & \xmark & \cmark & \xmark & \cmark & \cmark & \xmark & &  &  \\
         {\jlthirdred PEAC$_{(ad,g)}^{-2}$} & \xmark & \cmark & \cmark & \cmark & \cmark & \xmark &  &  &  \\ 
         \cdashline{2-10}
         & \xmark  & \cmark & \cmark & \xmark & \xmark & \cmark  &  &  &  \\
         & \xmark & \cmark & \xmark & \cmark & \xmark & \cmark  & &  &  \\
         {\jlthirdred PEAC$_{(ad,l)}^{-2}$} & \xmark & \cmark & \cmark & \cmark & \xmark & \cmark  &  &  &  \\ 
         \cdashline{2-10}
         & \xmark  & \cmark & \cmark & \xmark & \cmark & \cmark  &  &  &  \\
         & \xmark & \cmark & \xmark & \cmark & \cmark & \cmark  & &  &  \\
         {\jlthirdred PEAC$_{ad}^{-2}$($\equiv$PEAC$_{(ad,g,l)}^{-2}$)} & \xmark & \cmark & \cmark & \cmark & \cmark & \cmark  &  &  &  \\         
    \midrule
         \jlthirdreplace{PEAC$_{(o,a,g)}^{-2}$}{PEAC$_{(od,ad,g)}^{-2}$}& \cmark & \cmark & \cmark & \cmark & \cmark & \xmark & $81.25 \pm 0.16$ & $96.91 \pm 0.07$ & $73.35 \pm 0.19$\\
         & \cmark & \cmark & \cmark & \cmark & \xmark & \cmark &   &  & \\    
         PEAC$^{-2}$\jlthirdinsert{($\equiv$PEAC$_{od,ad,g,l}^{-2}$)}& \cmark & \cmark & \cmark & \cmark & \cmark & \cmark & $81.38 \pm 0.03$ & $97.14 \pm 0.10$ & $74.19 \pm 0.15$ \\         
    \bottomrule
    \end{tabular}
\end{table}
}

\section{Ablations: Local and global consistency}


\subsection{\jlthirdreplace{Local}{PEAC local} consistency improves performance}


We add the local consistency loss based on several methods VICRegL~\cite{bardes2022vicregl}, SimMIM~\cite{xie2022simmim} shown in \tablename\ref{tab:local consistency}. In the instance of VICRegL, ConvNeXt serves as the backbone, with the subsequent addition of local consistency loss precipitating notable enhancements in performance across all three target tasks. The SimMIM methodology employs Swin-B as its backbone, with the sequential addition of global and local consistency losses leading to marked improvements in performance.
Moreover, the removal of local consistency loss from our PEAC method corresponds to a decline in performance across the target classification tasks. This evidence underscores the efficacy of our proposed grid-matched local consistency loss.

\begin{figure}[h!]
    \centering
    \setlength{\belowcaptionskip}{0.1cm}
    \includegraphics[width=12cm]{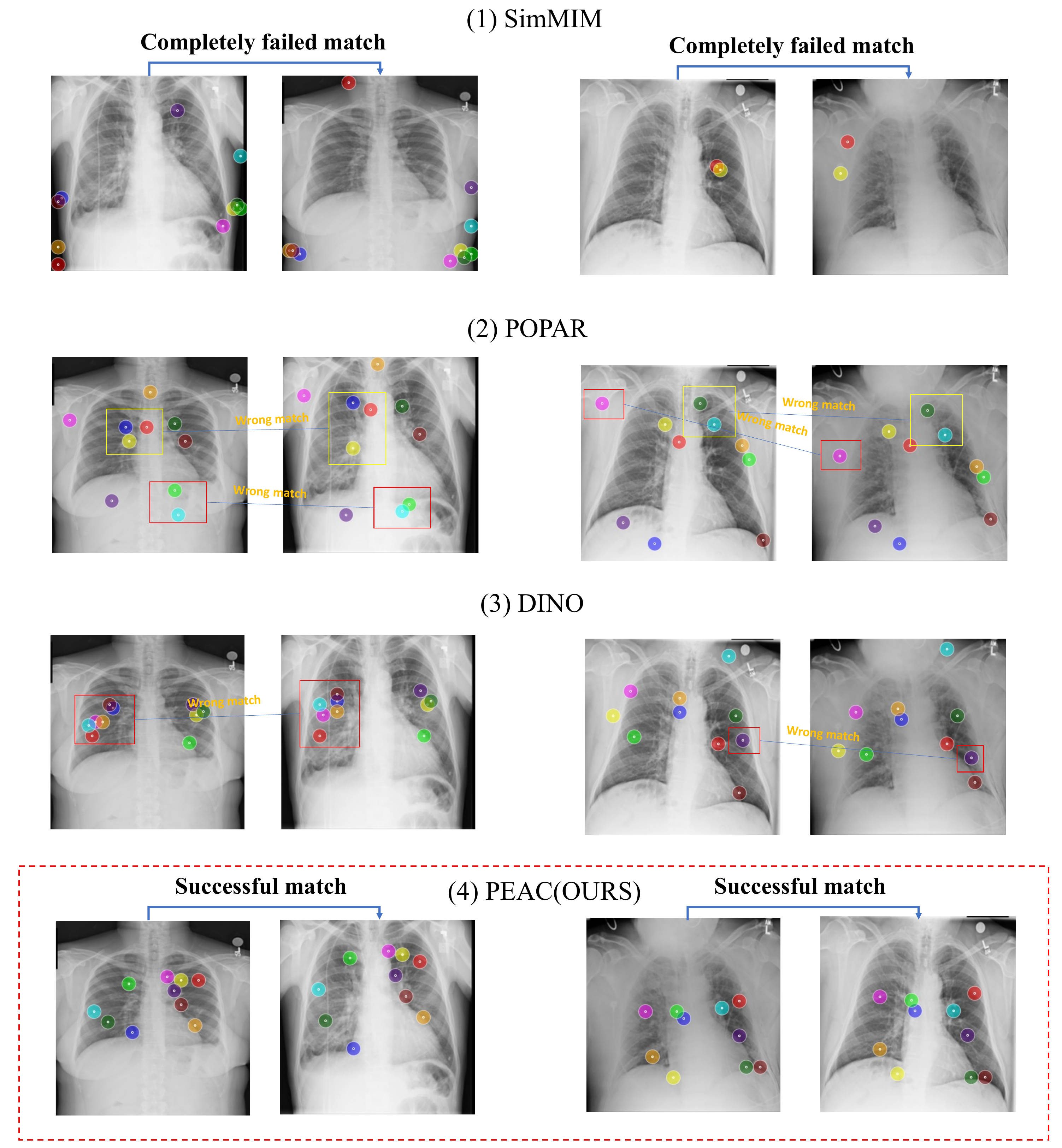}
    \caption{\jlthirdnote{\ul{I would move the methods: DINO, POPAR, SimMIM, and PEAC at the top and bottom of each subfigure.}}{} \jlthirdnote{It might not be fair to compare PEAC with DINO, POPAR, and SimMIM. You may want to focus comparisons with those methods for regional contrast learning and for dense embedding.}{} \jlthirdreplace{The correspondence results for cross different patients comparing with other methods. For the same pair of patient images, our PEAC achieves the best local anatomy matching compare with other methods. The correspondence of SimMIM is of completely failed match, POPAR matches some local patches while others are wrong or imprecise, and DINO gets better performance than SimMIM and POPAR but some patches are not correct.}{Comparing PEAC with DINO, POPAR, and SimMIM in matching anatomical strictures across distinct patients. For the same pair of patient images, our PEAC provides the most reliable anatomy matching. \jlthirdnote{Please develop a quantitative measure to evaluate the whole test dataset in comparison with DINO, POPAR, and SimMIM; better yet with those methods for regional contrast learning and for dense embedding.}{}}}
    \label{fig:cross-patient_correspondence}
\end{figure}

\subsection{\jlthirdreplace{Teacher-Student model (global consistency) boosts all one branch methods}{PEAC global consistency boosts performance}}

Corresponding to the main paper in \jlthirdreplace{chapter}{Section} 4.3 (3) our experiments in \tablename\ref{tab:teacher-student} demonstrate that using Teacher-Student model with global embedding consistency can boost one branch methods. We conduct experiments based on SimMIM and our own method which are all based on Swin-B backbone, pretrained on ChestX-ray14~\cite{wang2017chestx}, pretrained and fine-tuned on 224 image resolution. When adding teacher branch for SimMIM to compute the global embedding consistency loss, the classification performances of SimMIM$_{(g)}$ for the three target tasks are significantly improved. Importantly, the input images of the two branches are the two global views which are grid-wise cropped using our method and the student branch in SimMIM$_{(g)}$ gets the masked patches as SimMIM while the teacher branch gets no augmentations for the input images. We also add teacher branch to the one branch methods POPAR$_{od}^{-2}$ and POPAR$^{-2}$ for computing the global consistency loss. The downstream performances on the two branches Teacher-Student models PEAC$_{(o,g)}^{-2}$ and PEAC$_{(o,a,g)}^{-2}$ are much better than one branch methods POPAR$_{od}^{-2}$ and POPAR$^{-2}$.

\begin{figure}[h!]
    \centering
    \setlength{\belowcaptionskip}{0.1cm}
    \includegraphics[width=13cm]{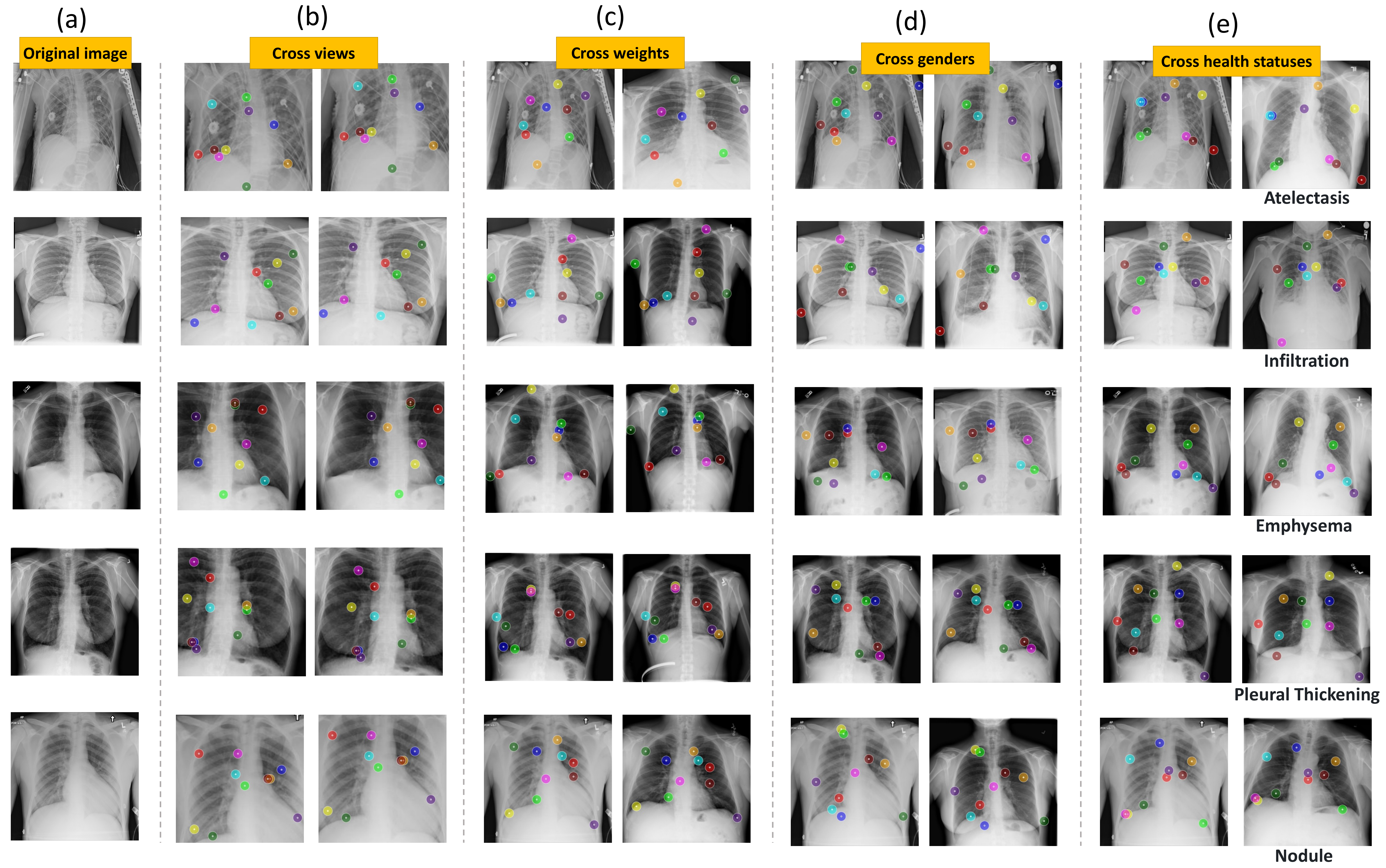}
    \caption{\jlthirdnote{Actually, what I was suggesting is that you have a reference image with a set of randomly generated fixed points and match this set of points across chest X-rays of the opposite gender, different weights, various health statuses, etc. Can you make a video on this? It is good enough for now to pre-compute all dense embeddings for those chest X-rays in the demo. You do not need to do this for this version but do so for the camera-ready version if accepted.}{} \jlthirdnote{\ul{too wordy}}{Establishing anatomical correspondence across views, across subject weights, across genders, and across health statuses. (a) original images of patients with no detected pulmonary disease. (b) cross-view correspondences, utilizing cropped images from the original set. (c) cross-weight correspondences among patients with significant differences in weight. (d) cross-gender correspondences between male and female patients. (e) correspondences among patients with varied health conditions. The left chest X-rays in (c), (d), and (e) are the same original images from (a), contrasting with the right ones that feature patients with distinct genders, weights, and health statuses. In (e), we also indicate diseases under the respective images.
    }}
    \label{fig:correspondence 5group}
\end{figure}

\section{Visualization of Upstream Models}

\subsection{Cross-patient and cross-view correspondence}

\begin{figure}[h!]
    \centering
    \setlength{\abovecaptionskip}{0.cm}
    \includegraphics[width=7cm]{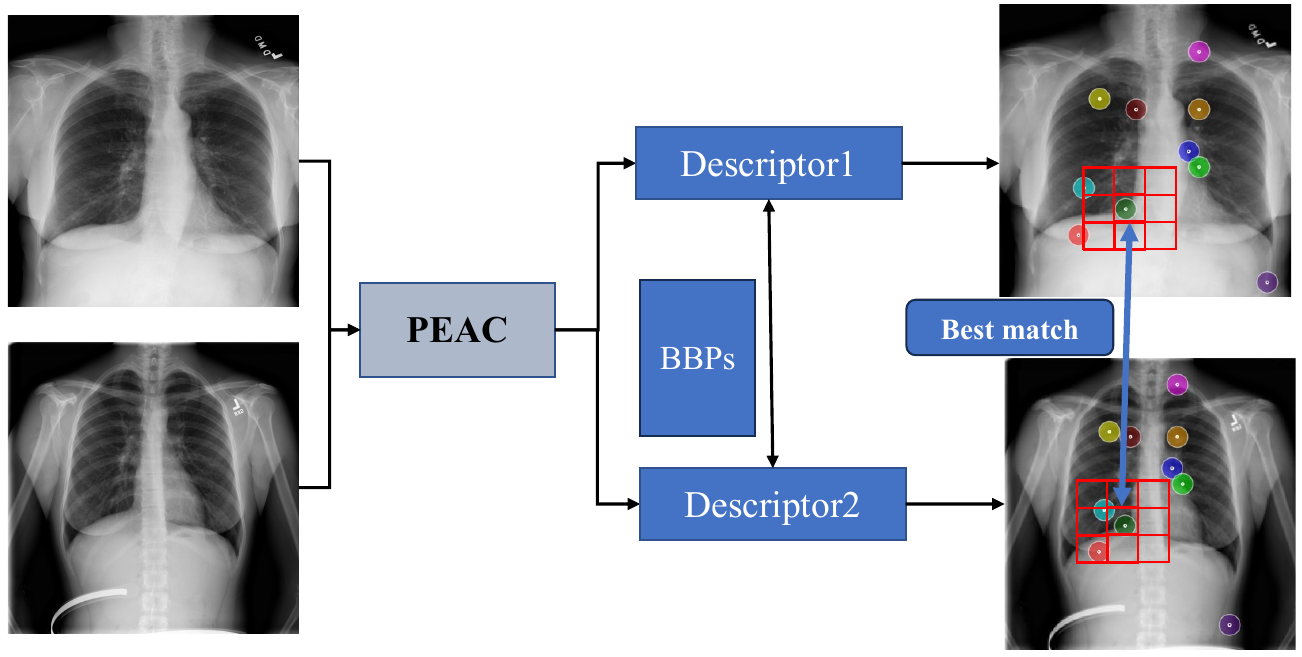}
    \jlcleandelete{\caption{ViT should be PEAC?? {\jlred I understand what is illustrated here, but what I did not understand is what is included within jlnote, that is ``their X-ray images are divided into $ 1+(1+(224-16)/4)^2=2810$ patches, where 16 is origin patch size of ViT-B/16 (PEAC$^{-3}$) to avoid oversteping the image boundaries, $4$ is new patch size for generate more detailed local description, first $1$ is for CLS Token, the second $1$ compensates for edge patch. and finally resulting in \jlnote{53x53=2809??}{$2810$} corresponding embedding vectors that represent the initial image patches. Then using average pooling with increasing kernal size on spatial dimension to generate multiple scales description.''}}}
    \caption{Pipeline of Corresponding patch match. After training, during inference, PEAC has only one single image as input. We eatablish cross-view or cross-patient correspondences by first computing dense embedding for each image. Specifically, we Generate 53×53 dense embeddings using a 14x14 grid with patch size of 16×16. The grid is shifted four times by 4 pixels to the right and for each of the shift to the right, it is further shifted four times by 4 pixels downward, leading to a grid of 53×53. Then we  calculate the cosine similarity metric of embedding vectors from one image to another and find N Best-Buddies Pairs (BBPs)~\cite{oron2017best}. Finally, from the BBPs we selected the top10 most similar image patch pairs with K-means and drew the correspondence points.
    }
    \label{fig:match_division} 
\end{figure}

To investigate the promotion of our method for sensing local anatomy, we match small local patches \jlthirdreplace{cross}{across} two patients' and one patient's different views of X-ray. \figname\ref{fig:cross-patient_correspondence} shows the cross-patient correspondence of our PEAC and other methods. \jlthirdnote{This is too coarse, and generating dense embedding is needed.}{Following ~\cite{amir2021deep} we divide each image with a resolution of 224 into 196 image patches using ViT-B backbone, and match the patch embedding of each image patch to the most similar patch embedding in another image. Finally, we selected the top 10 most similar image patches with K-means and drew the correspondence points.}  By comparing the correspondence results of our methods with SimMIM, POPAR and DINO in \figname\ref{fig:cross-patient_correspondence}, we learn that our method PEAC can learn the local anatomy more precisely.  The details of the algorithm are shown in  \figname\ref{fig:match_division}.

We also use our PEAC method to match anatomical structures from a patient with no finding (disease) to patients of different weights, different genders, and different health statuses as shown in \figname\ref{fig:correspondence 5group}. The results show that our PEAC can \jlthirdreplace{stably}{consistently} and precisely capture similar anatomies across different views of the same patients and across patients of opposite genders, different weights, and various health statuses.

\end{document}